\documentclass[lettersize,journal]{IEEEtran}
\usepackage{amsmath,amsfonts}
\usepackage{algorithmic}
\usepackage{algorithm}
\usepackage{array}
\usepackage[caption=false,font=normalsize,labelfont=sf,textfont=sf]{subfig}
\usepackage{textcomp}
\usepackage{stfloats}
\usepackage{url}
\usepackage{verbatim}
\usepackage{graphicx}
\usepackage{cite}
\usepackage[pagebackref,breaklinks,colorlinks]{hyperref}
\usepackage{bm}
\usepackage{multirow}
\usepackage{booktabs}
\usepackage{algorithm}
\usepackage{algorithmic}

\usepackage{xcolor}
\usepackage{colortbl} 

\begin{document}

\title{Beyond Inserting: Learning Identity Embedding for Semantic-Fidelity Personalized Diffusion Generation}

\author{Yang Li$^*$,
        Songlin~Yang\textsuperscript{*},~\IEEEmembership{Student Member,~IEEE,}
        \\Wei~Wang\textsuperscript{\dag},~\IEEEmembership{Member,~IEEE,}
        and~Jing~Dong,~\IEEEmembership{Senior Member,~IEEE}%
        \thanks{$^*$ indicates equal contribution.}
        \thanks{\dag indicates corresponding author.}
        }


\markboth{Submitted to IEEE TRANSACTIONS ON CIRCUITS AND SYSTEMS FOR VIDEO TECHNOLOGY}%
{Shell \MakeLowercase{\textit{et al.}}: A Sample Article Using IEEEtran.cls for IEEE Journals}



\twocolumn[{
\renewcommand\twocolumn[1][]{#1}
\maketitle
    \begin{figure}[H]
    \vspace{-0.8cm}
    \hsize=\textwidth 
    \centering
    \includegraphics[scale=0.57]{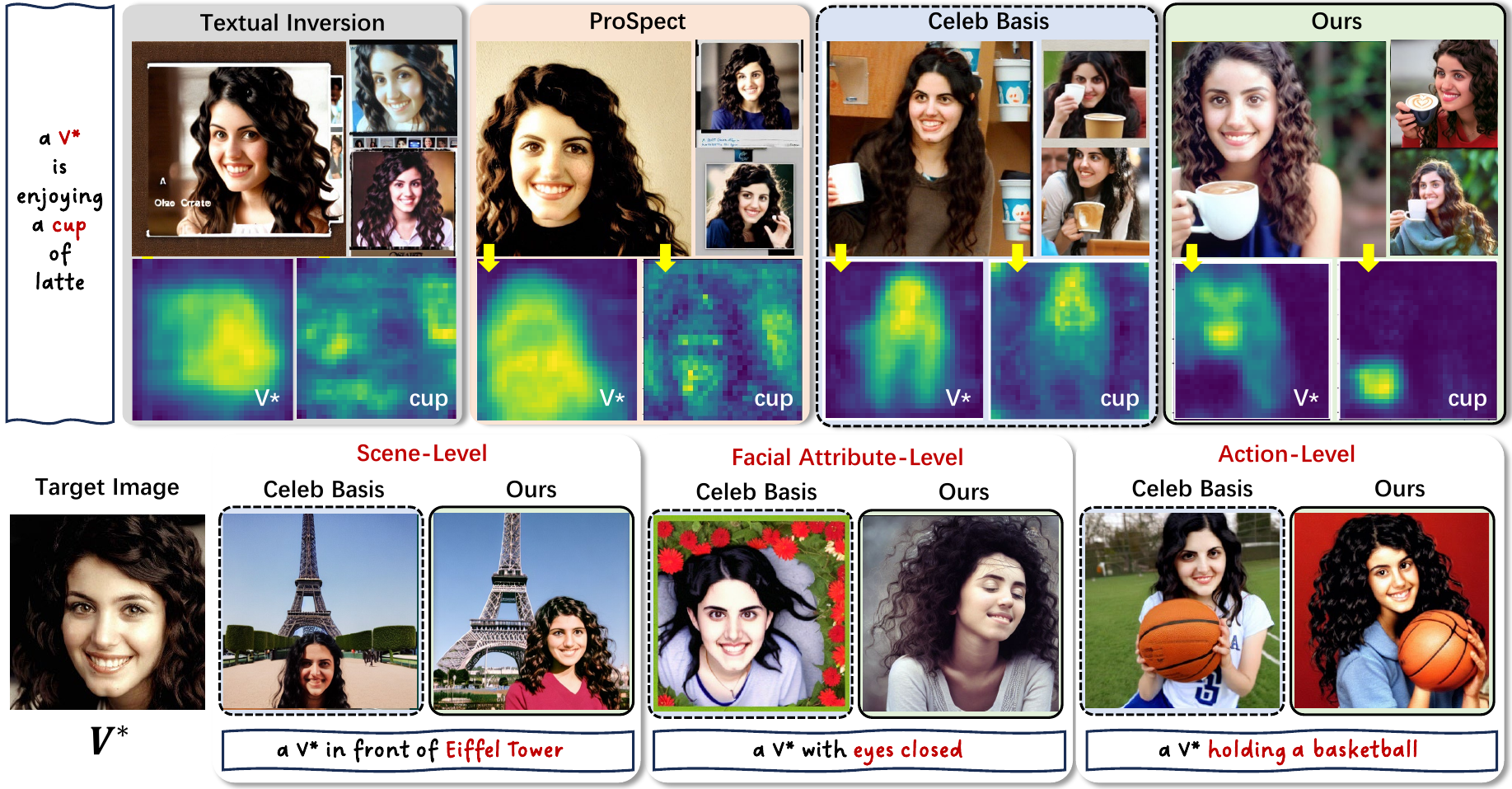}
    \caption{Previous methods for inserting new identities (IDs) into pre-trained Text-to-Image diffusion models for personalized generation have two problems: \textcolor{purple}{\textit{\textbf{(1) Attention Overfit}}} : As shown in the activation maps of Textural Inversion~\cite{gal2022image} and ProSpect~\cite{zhang2023prospect}, their ``V*'' attention nearly takes over the whole images, which means the learned embeddings try to encode both the human faces and ID-unrelated information in the reference images, such as the face region layout and background. This problem extremely limits their generative ability and disrupts their interaction with other existing concepts such as ``cup'', which results in the failure of the given prompt (i.e., they fail to generate the image content aligned with the given prompt). \textcolor{purple}{\textit{\textbf{(2) Limited Semantic-Fidelity}}}: Despite alleviating overfit, Celeb Basis~\cite{yuan2023inserting} introduces excessive face prior, limiting the semantic-fidelity of the learned ID embedding (e.g., the ``cup'' attention still continues to the ``V*'' face region and this limitation hinders the control of facial attributes such as ``eyes closed''). Therefore, we propose \textbf{Face-Wise Region Fit (Sec.~\ref{attention loss})} and \textbf{Semantic-Fidelity Token Optimization (Sec.~\ref{k-v feature disentangle})} to address problem (1) and (2) respectively. \textbf{\textit{More results: \small{\href{https://com-vis.github.io/SeFi-IDE/}{https://com-vis.github.io/SeFi-IDE/}}.}}}
    \label{teaser}
    \end{figure}
}]

\input{parts/1_abs}


\section{Introduction}
\label{sec:introduction}


\IEEEPARstart{R}{ecently}, Text-to-Image (T2I) models, such as the Stable Diffusion Model~\cite{rombach2022high}, have demonstrated an impressive ability to generate diverse, high-quality, and semantic-fidelity images using text prompts alone, thanks to image-aligned language encoders~\cite{radford2021learning} and diffusion-based generative models~\cite{dhariwal2021diffusion,nichol2021improved}. However, the challenge of personalized generation still remains, because the accurate person-specific face manifold can not be represented by text tokens, especially for the non-famous users whose data are not included in the training dataset. \textbf{In this paper, we focus on learning the accurate identity embedding for semantic-fidelity personalized diffusion-based generation using only one face image.}

The previous methods for this task have two problems that need to be addressed: \textit{\textbf{(1) Attention Overfit:}} Their fine-tuning strategies~\cite{kumari2023multi,ruiz2023dreambooth}, such as Texural Inversion~\cite{gal2022image} and ProSpect~\cite{zhang2023prospect}, tend to fit the whole target image rather than the ID-related face region, which entangle face layout and background information into the ID embedding. This results in the low ID accuracy and the difficulty of generating other existing concepts in the given prompt, such as ID-unrelated scenes (e.g., ``Eiffel Tower''), ID-related facial attributes (e.g., expressions and age), and actions (e.g., ``holding a basketball''). Particularly for actions, it is more challenging to generate prompt-fidelity human motions and human-object interactions, which can be shown in Fig.~\ref{teaser}. \textit{\textbf{(2) Limited Semantic-Fidelity:}} Their ID embedding methods lack the semantic-fidelity representations for facial attributes, which results in that human faces are treated as objects without non-rigid and diverse deformations. Although Celeb Basis~\cite{yuan2023inserting} can achieve an accurate ID mapping, it is unable to manipulate the facial attributes of the target image, such as expressions (e.g., ``eyes closed'' in Fig.~\ref{teaser}).

To address these problems, we propose our identity embedding method from two perspectives: \textit{\textbf{(1) Face-Wise Region Fit}}: We first visualize the attention overfit problem of the previous methods from the attention feature activation maps and then propose a face-wise attention loss to fit the face region instead of the whole target image. This key trick can improve the ID accuracy and interactive generative ability with the existing concepts in the original Stable Diffusion Model. \textit{\textbf{(2) Semantic-Fidelity Token Optimization}}: We optimize one ID representation as several per-stage tokens, and each token consists of two disentangled features. This approach expands the textual conditioning space and allows for semantic-fidelity control ability. Our extensive experiments validate that our method achieves higher accuracy in ID embedding and is able to produce a wider range of scenes, facial attributes, and actions compared to previous methods.

\textbf{To summarize, the contributions of our approach are:}
\begin{itemize}
    \item We visualize attention overfit problem of the previous methods, and propose a face-wise attention loss for improving the ID embedding accuracy and interactive generative ability with the existing concepts in the original Stable Diffusion Model.
    \item For semantic-fidelity generation, we optimize one ID representation as several per-stage tokens with disentangled features, which expands the textual conditioning space of the diffusion model with control ability for various scenes, facial attributes, and actions.
    \item Extensive experiments validate our advantages in ID accuracy and manipulation ability over previous methods.
    \item Our method does not rely on any prior facial knowledge, which has the potential to be applied to other categories.
\end{itemize}


\begin{figure*}[!ht]
    \centering
    \includegraphics[scale=0.65]{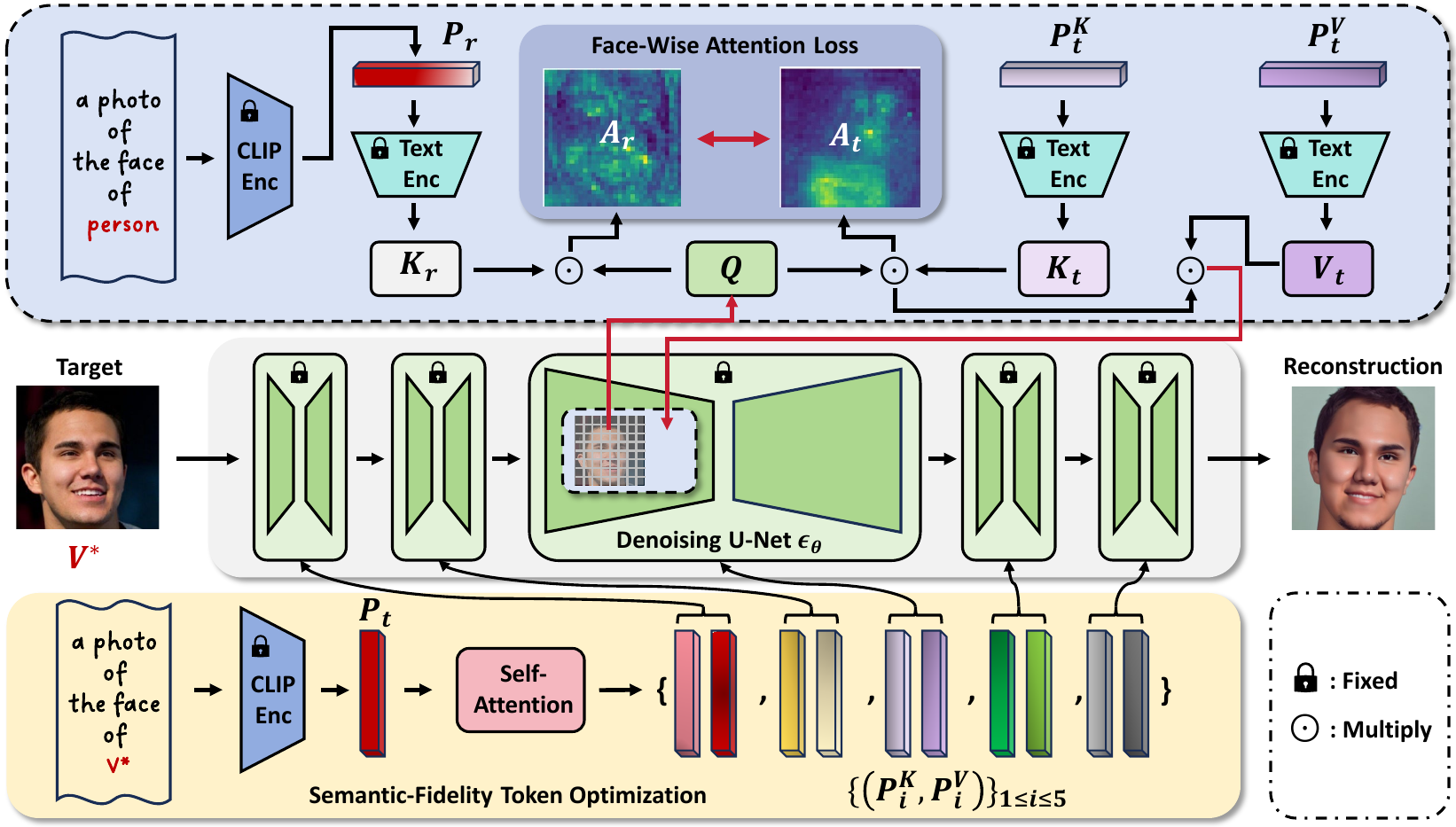}
    \vspace{-0.5cm}
    \caption{The overview of our framework. We first propose a novel \textbf{Face-Wise Attention Loss (Sec.~\ref{attention loss})} to alleviate the attention overfit problem and make the ID embedding focus on the face region to improve ID accuracy and interactive generative ability. Then, we optimize the target ID embedding as five per-stage tokens pairs with disentangled features to expend textural conditioning space with \textbf{semantic-fidelity control ability (Sec.~\ref{k-v feature disentangle})}. }
    \label{pipeline}
    \vspace{-0.4cm}
\end{figure*}

\section{Related Work}
\label{sec:related work}

\subsection{Text-Based Image Synthesis and Manipulation}
\label{text-driven editing}

Previous models such as GAN~\cite{karras2019style,yuan2019bridge,goodfellow2020generative,cheng2021rifegan2,yang2023designing,tan2023attention,yang2023context,han2023possible,yang2023learning,zuo2024statistics}, VAE~\cite{kingma2013auto,van2017neural,yang2023exposing,yang2021systematical}, Autoregressive~\cite{ramesh2022hierarchical,esser2021taming}, Flow~\cite{ramesh2021zero,dinh2014nice} were adopted to model the dataset distribution, and then synthesize new realistic images through sampling from the modeled distribution. Based on these, text-driven image manipulation~\cite{gal2022stylegan, patashnik2021styleclip, sun2024anyface++} has achieved significant progress using GANs by combining text representations such as CLIP~\cite{radford2021learning}. These methods work well on structured scenarios (e.g. human face editing), but their performance in fine-grained multi-modal alignment is not very satisfactory. Recent advanced diffusion models~\cite{dhariwal2021diffusion,nichol2021improved} have shown excellent diversity and fidelity in text-to-image synthesis~\cite{nichol2022glide, huang2022draw,ramesh2022hierarchical,rombach2022high,saharia2022photorealistic,chang2023muse}. Conditioned on the text embedding of the text encoder~\cite{radford2021learning}, these diffusion-based models are optimized by a simple denoising loss and can generate a new image by sampling Gaussian noise and a text prompt. Thanks to the powerful capabilities of diffusion in T2I generation, works~\cite{kim2022diffusionclip, wang2024unified, song2024doubly} achieve state-of-the-art text based image editing quality over diverse datasets, often surpassing GANs. Although most of these approaches enable various global or local editing of an input image, all of them have difficulties in generating novel concepts~\cite{chen2022re} or controlling the identity of generated objects~\cite{gal2022image}. Existing methods either directly blended the latent code of objects~\cite{avrahami2022blended, yang2023eliminating} to the generated background, or failed to understand the scenes correctly~\cite{yang2023paint}, which results in the obvious artifacts. To further solve this problem, some work~\cite{hertz2022prompt,mokady2023null,parmar2023zero,qi2023fatezero} adopted attention-based methods to manipulate target objects, but fail to balance the trade-off between content diversity and identity accuracy.

\subsection{Personalized Generation of Diffusion-Based T2I Models}
\label{personlized synthesis}

Using images from the new concepts for fine-tuning can obtain a personalized model, which can insert new concepts into the original model and synthesize concept-specific new scenes, appearances, and actions. Inspired by the GAN Inversion~\cite{xia2022gan}, recent diffusion-based personalized generation works can be divided into three categories: \textit{\textbf{(1) Fine-Tuning T2I model}}: DreamBooth~\cite{ruiz2023dreambooth} fine-tunes all weight of the T2I model on a set of images with the same ID and marks it as the specific token. \textit{\textbf{(2) Token Optimization}}: Textual Inversion~\cite{gal2022image}, ProSpect~\cite{zhang2023prospect}, and Celeb Basis~\cite{yuan2023inserting} optimize the text embedding of special tokens to map the specific ID into the T2I model, where the T2I model is fixed in the optimization process. \textit{\textbf{(3) Tuning Free}}: ELITE~\cite{Wei_2023_ICCV} learning an encoder to customize a visual concept provided by the user without further fine-tuning. BootPIG~\cite{purushwalkam2024bootpig} follows a bootstraping strategy by utilizing a pre-trained U-Net model to steer the personalization generation. Except for those, Token Optimization and Fine-Tuning are combined to manipulate multi-concept interactions~\cite{kumari2023multi} or saving fine-tuning time and parameter amount~\cite{smith2023continual,tewel2023key, arar2023domain, chen2024disendreamer}. 

\noindent
\textbf{ID Embedding for Faces.} Previous methods~\cite{shi2023instantbooth,gal2023designing} try to train an inversion encoder for face embedding, but face ID-oriented mapping is difficult to be obtained from a naively optimized encoder. Moreover, fine-tuning the T2I model on large-scale images often causes concept forgetting. For this, Celeb Basis~\cite{yuan2023inserting} adopts a pre-trained face recognition model and a face ID basis to obtain an ID representation for one single face image, and Face0~\cite{valevski2023face0} learned to project the embeddings of recognition models to the context space of Stable Diffusion. Except for ID representation, FaceStudio~\cite{yan2023facestudio} deployed a CLIP vision encoder~\cite{radford2021learning} to extract the structure features. InstantID~\cite{wang2024instantid} handled image generation in various styles by designing a learnable IdentityNet to grasp strong semantics.  However, introducing too strong face prior makes it difficult to manipulate diverse facial attributes and fails to generalize to other concept embedding. FastComposer~\cite{xiao2023fastcomposer} used a delayed subject conditioning strategy to avoid subject overfitting, but they only focus on faces and fail to interact with other objects such as ``sofa'' as shown in Fig.~\ref{action}. While PhotoMaker~\cite{li2023photomaker} proposed an ID-oriented dataset that includes diverse scenarios and fine-tuning part of the Transformer~\cite{vaswani2017attention} layers in the image encoder to mitigate contextual information loss. Nevertheless, the training of Transfromer will sacrifice the compatibility with existing pretrained community models.

\section{Method}
\label{sec:method}

Embedding one new identity (ID) into the Stable Diffusion Model for personalized generation using only \textbf{one single face image} has three technical requirements: accuracy, interactivity, and semantic-fidelity. Our learned ID embedding focuses on the face region and adopts disentangled token representation, which has flexible face spatial layout, interactive generation ability with existing concepts (e.g., generating interaction motion with other objects), and fine-grained manipulation ability (e.g., editing the facial expressions). This means that our method improves both ID accuracy and manipulation ability. As shown in Fig.~\ref{pipeline}, we propose our ID embedding pipeline from two key perspectives: \textit{\textbf{(1) Face-Wise Attention Loss}}: Towards the improvements in ID accuracy and interactive generative ability with existing concepts in the original model, we propose a face-wise attention loss in Sec.~\ref{attention loss}. \textit{\textbf{(2) Semantic-Fidelity Token Optimization}}: For diverse manipulation, we optimize one ID representation as several per-stage tokens, and each token consists of two disentangled embeddings, which can be seen in Sec.~\ref{k-v feature disentangle}. In the following sections, we first give an introduction of the pre-trained Stable Diffusion Model~\cite{rombach2022high}, and  we then provide the details of our method.

\subsection{Preliminary}
\label{preliminaries}

\noindent
\textbf{Diffusion-Based T2I Generation.} Our utilized Stable Diffusion Model~\cite{rombach2022high} consists of a CLIP text encoder~\cite{radford2021learning}, an AutoVAE~\cite{razavi2019generating} and a latent U-Net~\cite{ho2020denoising} module. Given an image $\bm{x}\in \mathbb{R}^{H\times W\times 3}$ ($H$ and $W$ represent the size of target image), the VAE encoder $\mathcal{E}(\cdot)$ maps it into a lower dimensional latent space as $\bm{z}=\mathcal{E}(\bm{x})\in{\mathbb{R}^{h\times w\times c}}$ followed by a corresponding decoder $\mathcal{D}(\cdot)$ to map the latent vectors back as $\mathcal{D}(\mathcal{E}(\bm{x}))\approx \bm{x}$. The $h, w$ and $c$ are the dimensions of latent tensor $\bm{z}$. Given any user provided prompts $y$, the tokenizer of the CLIP text encoder $\mathcal{E}_{text}(\cdot)$ divides and encodes $y$ into integer tokens. Correspondingly, by looking up the dictionary, a word embedding group $\bm{Y}$ can be obtained. Then, the CLIP text transformers $\mathcal{T}_{text}$ encode $\bm{Y}$ to generate text condition vectors $\mathcal{T}_{text}(\bm{Y})$, which serve as a condition to guide the training of the latent U-Net denoiser $\epsilon_{\theta}(\cdot)$:
\begin{equation}
    \mathcal{L}=\mathbb{E}_{\bm{z}\sim \mathcal{E}(\bm{x}), \bm{\epsilon}\sim\mathcal{N}(0, 1), t}\left [ \left \| \epsilon - \epsilon_{\theta}(\bm{z}_t, t, \mathcal{T}_{text}(\bm{Y}))\right \| \right ],
\end{equation}
where $\bm{\epsilon}$ denotes for the unscaled noise and $t$ is the timestep. $\bm{z}_t$ is the latent variable of a forward Markov chain $\mathrm{s.t.} \bm{z}_{t}=\sqrt{\alpha_{t}}z_{0} + \sqrt{1 - \alpha_{t}}\bm{\epsilon}$, where $\alpha\in \left [0, 1\right ]$ is a hyper-parameter that modulates the quantity of noise added. Given a latent noise vector $\bm{z}_t$ in the timestep $t$, the model learns to denoise it to $\bm{z}_{t - 1}$. During inference, a random Gaussian latent vector $\bm{z}_T$ is iteratively denoised to $\bm{z}_0$. 

\begin{figure}[t]
    \centering
    \includegraphics[scale=0.45]{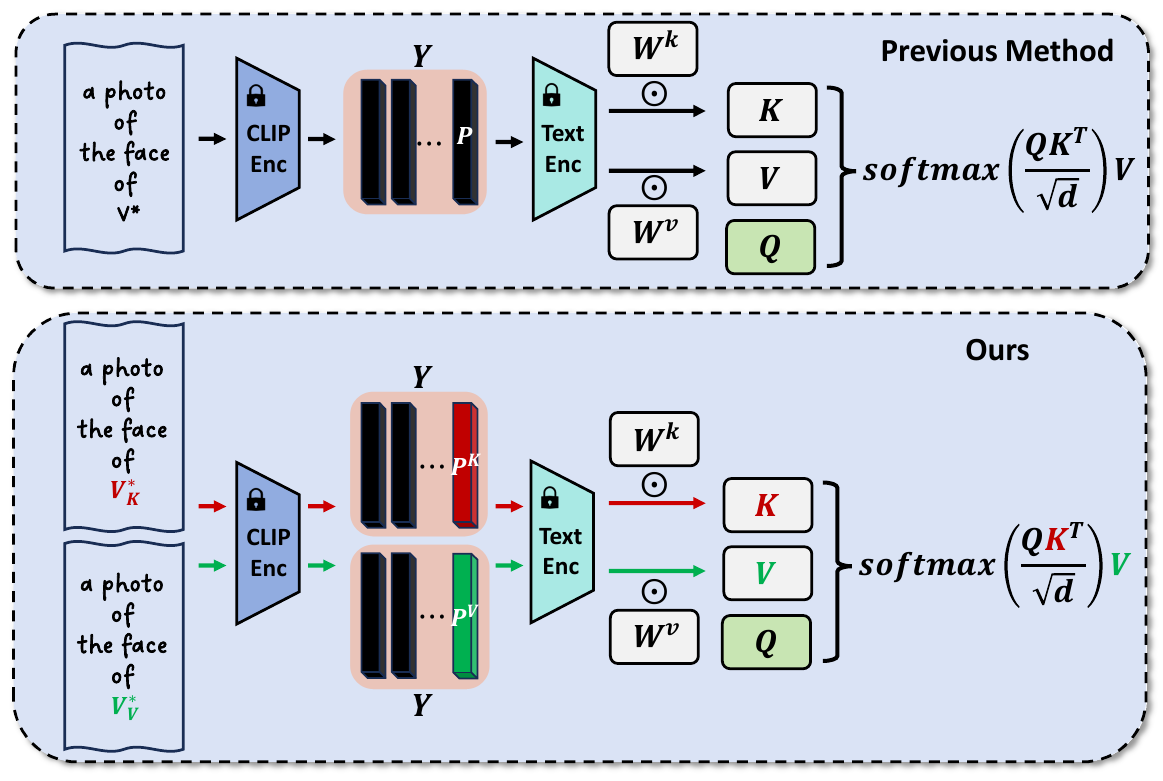}
    \caption{The details of text condition and K-V feature implementation differences.}
    \label{KV-split}
\end{figure}

\noindent
\textbf{Cross-Attention for Text Condition.} As shown in the upper block of Fig.~\ref{KV-split}, the text prompt is first tokenized to a word embedding group $\bm{Y}$, and then encoded by the text transformers to generate text condition $\mathcal{T}_{text}(\bm{Y})$. Given the latent image features $\bm{f}$, the cross attention operation updates the latent features as:
\begin{equation}
    \bm{Q} = \bm{W^q}\bm{f}, \bm{K} = \bm{W^k}\mathcal{T}_{text}(\bm{Y}), \bm{V} = \bm{W^v}\mathcal{T}_{text}(\bm{Y}), 
\end{equation}
\vspace{-0.5cm}
\begin{equation}
    Attention(\bm{Q},\bm{K}, \bm{V})=Softmax(\frac{\bm{Q}\bm{K^T}}{\sqrt{d}})\bm{V}.
\end{equation}
where $\bm{W^q}$, $\bm{W^{k}}$, and $\bm{W^v}$ map the inputs to \textbf{Q}uery, \textbf{K}ey, and \textbf{V}alue features, respectively. The $d$ is the output dimension of \textbf{K}ey and \textbf{Q}uery features.

Previous work has shown that the CLIP text embedding space is expressive enough to capture image semantics~\cite{gal2022image, zhang2023prospect}. Specifically, a placeholder string, ``$V^*$'', is designated in the prompt $y$ to represent the identity-related feature we wish to learn. During the word embedding process, the vector associated with the word ``$V^*$'' is replaced by the learned ID embedding $\bm{P}$. Thus, we can combine ``$V^*$'' with other words to achieve personalized creation. \textcolor{purple}{\textbf{In this work, we focus on learning accurate and interactive ID embedding $\bm{P}$}.}

\begin{algorithm}[t]
    \caption{Calculating the Face-Wise Attention Loss}
    \label{alg: attention loss}
    \begin{algorithmic}
    \REQUIRE Query features $\mathbf{Q}$, reference key features $\bm{K_r}$, target 
    \STATE key features $\bm{K_t}$, 
    \text{attention score modules \textbf{ATT} in cross- }
    \STATE attention blocks, and \text{attention map rearrange and resize} 
    \STATE \text{function \textbf{Re2}}.
    \ENSURE $\mathcal{L}_{attention}$
    \FOR{$i\leftarrow1$ to $S$}
        \STATE $\bm{A_r}.\text{append}(\mathbf{\textbf{ATT}_{i}}(\bm{Q}, \bm{K_r}))$
        \STATE $\bm{A_t}.\text{append}(\mathbf{\textbf{ATT}_{i}}(\mathbf{Q}, \bm{K_t}))$
    \ENDFOR
    \STATE Rearrange and resize the attention map before calculating the loss
    \STATE $ \bm{A_r} =\textbf{Re2}(\bm{A_r}) \in \mathbb{R}^{8 \times 77 \times{32} \times{32}}$
    \STATE $ \bm{A_t} = \textbf{Re2}(\bm{A_t}) \in \mathbb{R}^{8 \times 77 \times{32} \times{32}}$
    \STATE Calculate the Face-Wise Attention Loss
    \STATE $ \mathcal{L}_{attention} = \text{MSE}(\bm{A_t}, \bm{A_t})$
    \RETURN $\mathcal{L}_{attention}$
    \end{algorithmic}
\end{algorithm}

\subsection{Face-Wise Attention Loss}
\label{attention loss}
We first analyze and visualize the attention overfit problem of previous methods. Then, we present an accessible prior from the Stable Diffusion Model, instead of the face prior from other models such as face recognition models, to improve both the embedding accuracy and interactive generative ability at the same time. These are our motivations to propose our Face-Wise Attention Loss. Finally, we present the details of the loss implementations.

\begin{figure*}[t]
    \centering
    \includegraphics[scale=0.8]{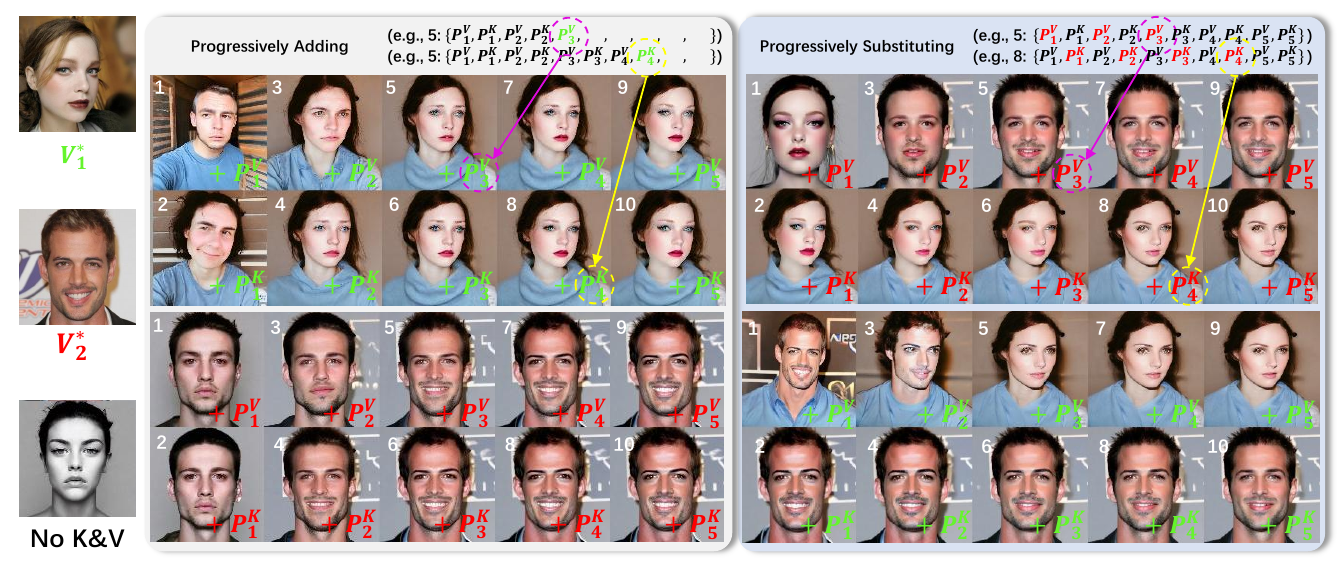}
    \caption{The different effects of $\bm{P_{i}^{K}}$ and $\bm{P_{i}^{V}}$ tokens. (1) Progressively Adding: We add different ${\{(\bm{P_i^{K}}, \bm{P_i^{V})}\}}_{1\leq i \leq 5}$ tokens to the conditioning information in ten steps. We found that the initial tokens effect more the layout of generation content (e.g., face region location, and poses), while the latter tokens effect more the ID-related details. (2) Progressively Substituting: We then substitute different $\bm{P_{i}^{K}}$ and $\bm{P_{i}^{V}}$ tokens of ${\{(\bm{P_i^{K}}, \bm{P_i^{V})}\}}_{1\leq i \leq 5}$. We found that $\bm{P_{i}^{V}}$ contribute to the vast majority of ID-related conditioning information, and the $\bm{P_{i}^{K}}$ contribute more to textural details, such as environment lighting.}
    \label{kv_interpolation}
\end{figure*}

\noindent
\textbf{Image Fit vs. Face-Region Fit.} Previous methods rely on learning from multiple images of a target object to grasp relevant features, such as Textual Inversion~\cite{gal2022image} and DreamBooth~\cite{ruiz2023dreambooth}. However, when only a single image is available, they are prone to fitting to the whole target image (including ID-unrelated face layout and background information), and the learned embedding tends to influence regions beyond the face region during the cross-attention stages. As a result, they lack the interactive generative ability with the existing concepts in the original model~\cite{tewel2023key}. In other words, during the inference, the generated results from the personalized model may not be consistent to the text prompts. For example, as shown in Fig.~\ref{teaser}, the given prompt is ``a $V^*$ is enjoying a cup of latte'', but the methods with attention overfit problem fail to generate the ``cup'' content. The same problem can also be seen in Fig.~\ref{figure3}, which given some facial attributes such as ``old'' in the prompt, the diffusion-based generation process just fails. Our ID embedding optimization can focus on ID-related face regions and neglect the unrelated background features, which can simultaneously improve the ID accuracy and interactive generative ability.

\noindent
\textbf{Make Best of Stable Diffusion Model.} Multiple target images are necessary for previous methods to acquire concept-related embedding. These images allow users to use text prompts to manipulate different poses, appearances, and contextual variations of objects. One target image fails to achieve this generalization. However, Stable Diffusion Model has learnt a strong general concept prior for various sub-categories. For example, different human identities belong to the general concept ``person'', and different dog categories such as Corgi and Chihuahua belong to the general concept ``dog''. Therefore, it is reasonable to adopt this prior knowledge to achieve one-shot learning. To meet our more higher requirements, we aim to have flexibility to manipulate the ID-specific regions of final images. In other words, when we want to generate images corresponding to ``a photo of $V^*$ is playing guitar'', only handling portrait or face image generation is not enough for this prompt. Therefore, we adopt ``the face of person'' as our general concept prior for ID embedding, because when provided with the prompt ``a photo of the face of person'', Stable Diffusion Model can generate a face image of a person with a randomly assigned identity and constrain the region where the generated person appears in the final image. 

Specifically, we propose to use a reference prompt $y_r$ that remains consistent with the general concept of different IDs, which replaces the placeholder word (``$V^*$'') with ``person'' in prompts (i.e., $y_r:=$``a photo of the face of person''). Then, we use this attention map derived from $y_r$ as a constraint to restrict the attention corresponding to the placeholder word (``$V^*$'') in the target prompt ``a photo of the face of $V^*$''. This approach allows the ID embedding to focus on the face region associated with the target ID while maintaining the coherence of the general concept. Specifically, we first embed the reference prompt and target prompt as word embedding groups $\bm{Y_r}$ and $\bm{Y_t}$. The ID embedding $\bm{P}$ in $\bm{Y_t}$ is fed into a self-attention module to obtain per-stage token embeddings ${\{(\bm{P_i^{K}}, \bm{P_i^{V})}\}}_{1\leq i \leq 5}$, which will be introduced in the subsequent section. Then, we adopt text encoder transformers $\mathcal{T}_{text}$ to obtain their corresponding key ($\bm{K}$) and value ($\bm{V}$) features ${\bm{K_{r}}, \bm{V_{r}}}$ and ${\bm{K_{t}}, \bm{V_{t}}}$. Then, each of the K features are send to the cross-attention module to calculate the attention map $\bm{A_r}$ and $\bm{A_t}$ with latent image features $\bm{Q}$ respectively. The $\bm{A_t}$ is constrained by the $\bm{A_r}$ within the corresponding representation of the concept as follows:
\begin{equation}
    \mathcal{L}_{attention} = ||\bm{A_r} - \bm{A_t}||_{2}^{2}.
\end{equation}
The detailed Face-Wise Attention Loss computation pipeline is depicted in Algorithm~\ref{alg: attention loss}.


\subsection{Semantic-Fidelity Token Optimization}
\label{k-v feature disentangle}
We first present the disadvantages of previous methods from the semantic-fidelity control. Then, we introduce our optimization strategy, including the motivation for feature disentanglement and the details of obtaining feature pairs. Finally, we present our training loss for optimization.

\noindent
\textbf{Lack of Semantic-Fidelity Control.} This problem can be found from two perspectives: \textit{\textbf{(1) Stable Diffusion Model}}: We observe that even though the face data of celebrities has been included in the training dataset of Stable Diffusion Model, it still fails to achieve perfect semantic-fidelity control for these IDs. For example, ``a photo of an old Obama'' cannot generate the corresponding images. \textit{\textbf{(2) Previous Personalized Methods}}: Methods like Celeb Basis~\cite{yuan2023inserting}, Textural Inversion~\cite{gal2022image} and InstantID~\cite{wang2024instantid} mainly emphasize how to preserve the characteristics of the person and achieve global control over the generated images through text modifications. Although these methods are able to manipulate scenes or styles, they struggle to control fine-grained facial attributes of learned IDs, such as age and expressions. Prospect~\cite{zhang2023prospect} represents an image as a collection of textual token embeddings which could offer better disentanglement and controllability in editing images. However, When it comes to the generation of images with controllable facial attributes, as shown in Fig.~\ref{figure3}, it fails to generate examples like ``an old person''. We address this challenge by disentangling the $\bm{K}$ and $\bm{V}$ features, as explained in detail in the following section.

\noindent
\textbf{Disentanglement of $\bm{K}$ and $\bm{V}$ Features.} The text condition features ($\bm{K}$, $\bm{V}$) will be fed into cross-attention layers of U-Nets for conditioning the generated images. Previous methods~\cite{kumari2023multi,tewel2023key} differentiated the $\bm{K}$ and $\bm{V}$ features calculated from the same ID embedding $\bm{P}$ as position information and object texture features, which is not appropriate for manipulating facial attributes. 
Therefore, to further investigate the different effects of $\bm{K}$, $\bm{V}$ features for our task, we disentangle the ID embedding $\bm{P}$ as per-stage token embedding groups $\{(\bm{P_i^{K}}, \bm{P_i^{V})}\}_{1\leq i \leq 5}$, and then visualize the effects of these features in the image generation process. As shown in Fig.~\ref{kv_interpolation}, we found that the $\bm{P^V_i}$ embeddings are more ID-related, while the $\bm{P^K_i}$ embeddings are more related to environment factors such as lighting, mouth open and face texture. The disentangled optimization of ID embedding in $\bm{P^K_i}$ and $\bm{P^V_i}$ can further improve the ID accuracy and interactive generative ability with other concepts. As shown in Fig.~\ref{KV-split}, we illustrate the different implementations. 

\begin{figure}
    \centering
    \includegraphics[scale=0.52]{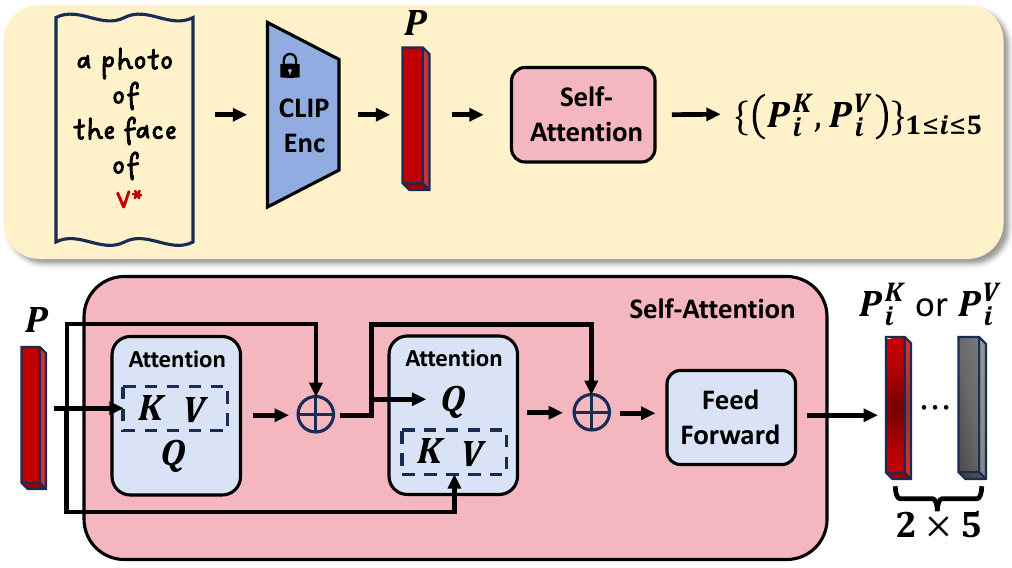}
    \caption{The details of Self-Attention module. For simplicity, we disregard the remaining embeddings in $\bm{Y_{t}}$ and focus on the ID embedding $\bm{P}$ associated with the pseudo-word ``V*''.}
    \label{self-attention}
\end{figure}

\noindent
\textbf{How to Obtain K-V Feature Pairs?} Specifically, the input prompt is firstly fed into the CLIP Tokenizer, which generates the textual token embeddings $\bm{Y_t}$. Here, the ID embedding $\bm{P}$ related to ``$V^*$'' is a vector with the size of $1\times 768$. As depicted in Fig.~\ref{self-attention}, the $\bm{P}$ is then fed to a trainable Self-Attention~\cite{vaswani2017attention} module to create $10 \times 768$ embedding $\{(\bm{P_i^{K}}, \bm{P_i^{V})}\}_{1\leq i \leq 5}$. Each of the newly generated ID embedding will replace the original ID embedding to form five groups of textual embeddings, and then these embedding groups will be multiplied by $\bm{W^k}$ and $\bm{W^v}$ to obtain $\bm{K}$ and $\bm{V}$ features. The Self-Attention module consists of two self-attention layers with one feed-forward layer. We take each group of textual embeddings as a different condition. We evenly divide the 1000 diffusion steps into five stages, each stage corresponds to a unique pair of textual embeddings. Finally, only the Self-Attention module is trainable and the final ${\{(\bm{P_i^{K}}, \bm{P_i^{V})}\}}_{1\leq i \leq 5}$ is obtained by optimizing the diffusion denoising loss as follows:
\begin{equation}
    \setlength\abovedisplayskip{1pt}
    \setlength\belowdisplayskip{1pt}
    \mathcal{L}_{K,V} = \mathbb{E}_{\bm{z}, t, \bm{P}}[||\epsilon - \epsilon_{\theta}(\bm{z_t}, t, (\bm{P_{i}^{K}}, \bm{P_{i}^{V}}))||_{2}^{2}],
\end{equation}
where $\bm{z}$ is the latent code of target image. The $\epsilon$ is the unscaled noise sample, and the $\epsilon_{\theta}$ is the U-Net module in diffusion model. The $t$ is the optimization step of the diffusion process, and the $\bm{z}_t$ is U-Net output of different steps.

\noindent
\textbf{Training Loss.} Our goal is seamlessly embedding one specific ID into the space of Stable Diffusion Model, which have to achieve accurate ID mapping and fully use the prior from the Stable Diffusion Model to manipulate scenes, actions and facial attributes. Thus, the total optimization objective can be formulated as follows:
\begin{equation}
    \vspace{-0.4cm}
    \mathcal{L} = \mathcal{L}_{K,V} + \lambda\mathcal{L}_{attention}.
\end{equation}

\begin{figure*}
    \centering
    \includegraphics[width=1.0\linewidth]{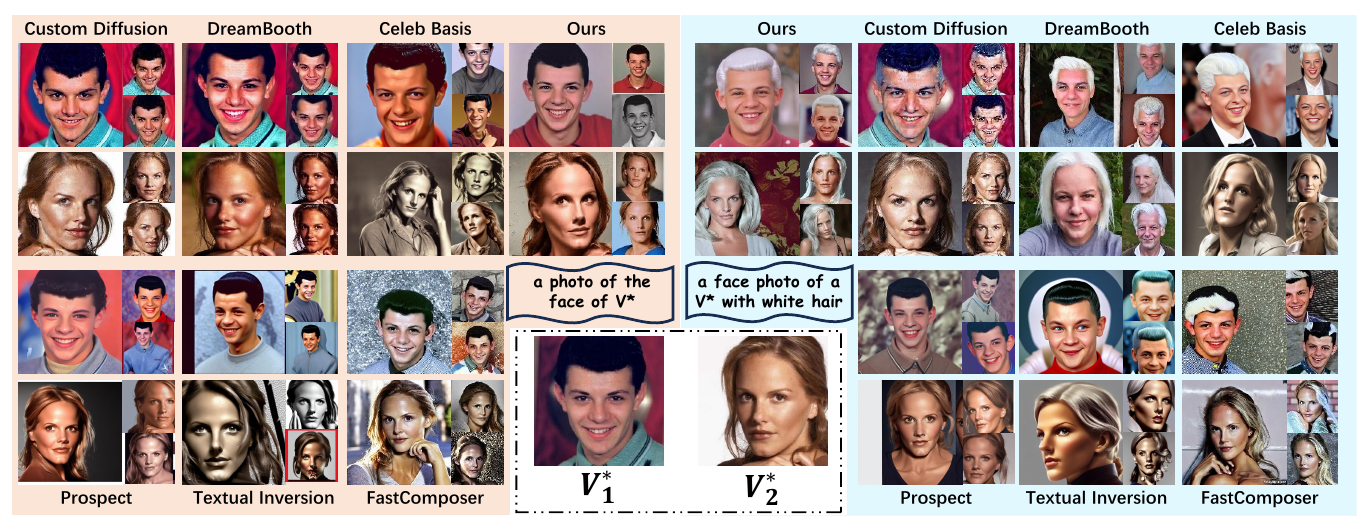}
    \caption{Face photo generation of ours and comparison methods. Due to attention overfitting, Textural Inversion~\cite{gal2022image} and Prospect~\cite{zhang2023prospect} struggle to generate images that accurately reflect the semantics of ``white hair``. Custom Diffusion~\cite{kumari2023multi} and DreamBooth~\cite{ruiz2023dreambooth} tend to overly mimic the training image and fail to maintain identity when combined with other text prompts. On the other hand, methods such as Celeb Basis~\cite{yuan2023inserting} and FastComposer~\cite{xiao2023fastcomposer} exhibit poor semantic fidelity and limited diversity in their generated outputs.}
    \label{face_generation}
\end{figure*}

\begin{figure*}[th!]
    \centering
    \includegraphics[scale=0.68]{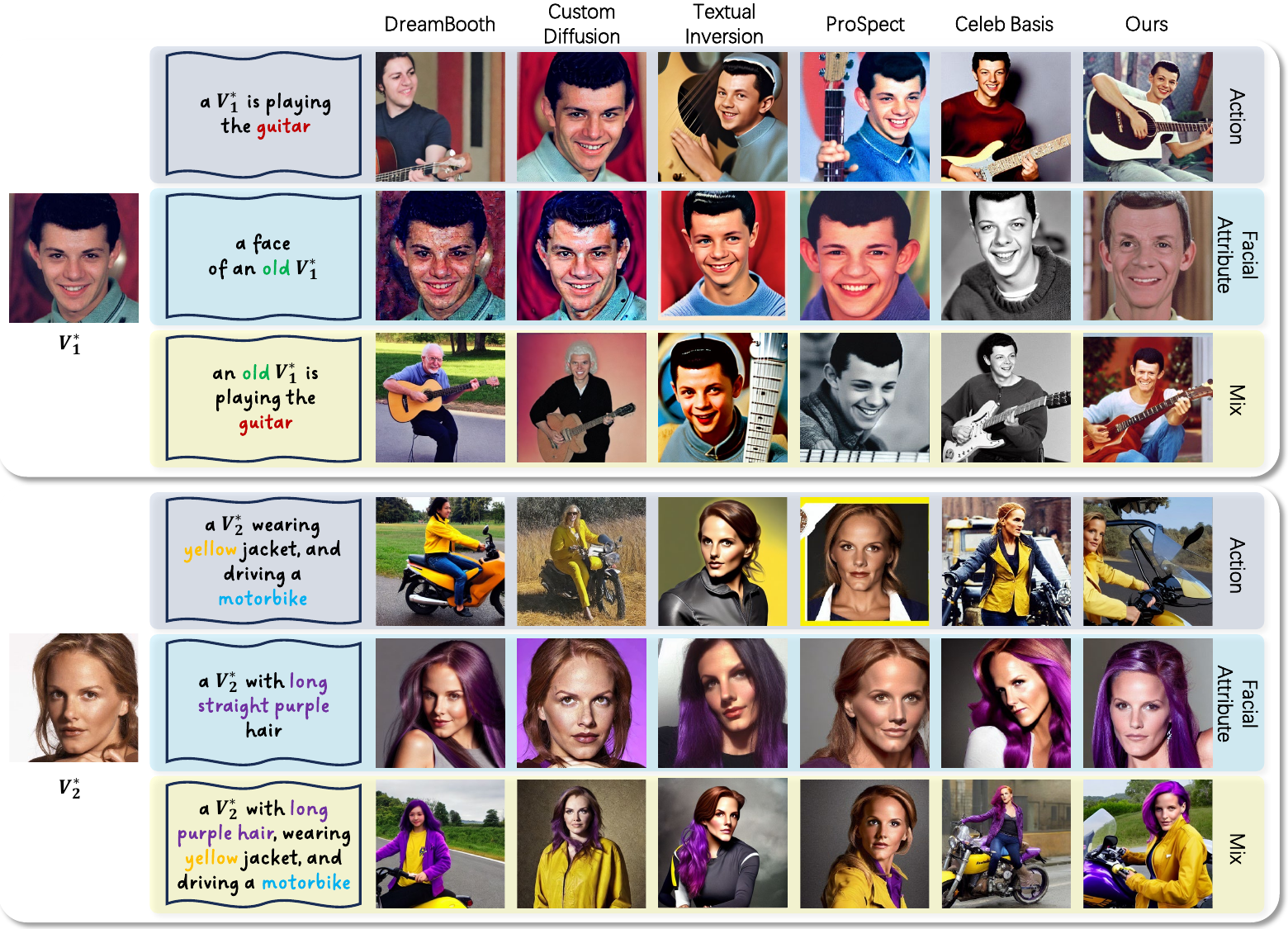}
    \caption{Qualitative comparisons with different SOTA methods using more complex prompts. We conduct experiments from three levels, including the action manipulation, facial attribute editing, and their mixture. Our method shows superior embedding accuracy and interactive generation ability with existing concepts.}
    \label{figure3}
\end{figure*}

\section{Experiments}
\label{sec:experiments}
\subsection{Experimental Settings}

\noindent
\textbf{Implementation Details.} We present our target T2I model, test data, training details, and inference recipe for reproductivity. \textit{\textbf{(1) Target T2I Model}}: Unless otherwise specified, we utilize Stable Diffusion 1.4~\cite{rombach2022high} with default hyper parameters as the pre-trained diffusion-based T2I model. We adopt a frozen CLIP model~\cite{saharia2022photorealistic} in the Stable Diffusion Model as the text encoder network. The texts are tokenized into start-token, end-token, and 75 non-text padding tokens. \textit{\textbf{(2) Test Data}}: The test face images are the StyleGAN~\cite{karras2020analyzing} synthetic data and the images from the CelebA-HQ dataset~\cite{karras2018progressive}. \textit{\textbf{(3) Training Details}}: For our method, the time for fine-tuning every ID using only one face image is $\sim 5$ minutes ($\sim 1000$ epochs) on one NVIDIA TITAN RTX GPU. We adopt Adam optimizer and set its learning rate as 0.005. The $\lambda$ is set to 0.003. Since we only rely on single face image to acquire its embedding, we adopt some image augmentation methods, including color jitter, horizontal flip with the probability of $0.5$, and random scaling ranging in $0.1 \sim 1.0$. \textit{\textbf{(4) Inference Recipe}}: During sampling time, we employ a DDIM sampler~\cite{song2020denoising} with diffusion steps $T=50$ and the classifier-guidance~\cite{ho2022classifier} with the guidance scale $w = 7.5$.

\begin{figure*}
    \centering
    \includegraphics[scale=0.62]{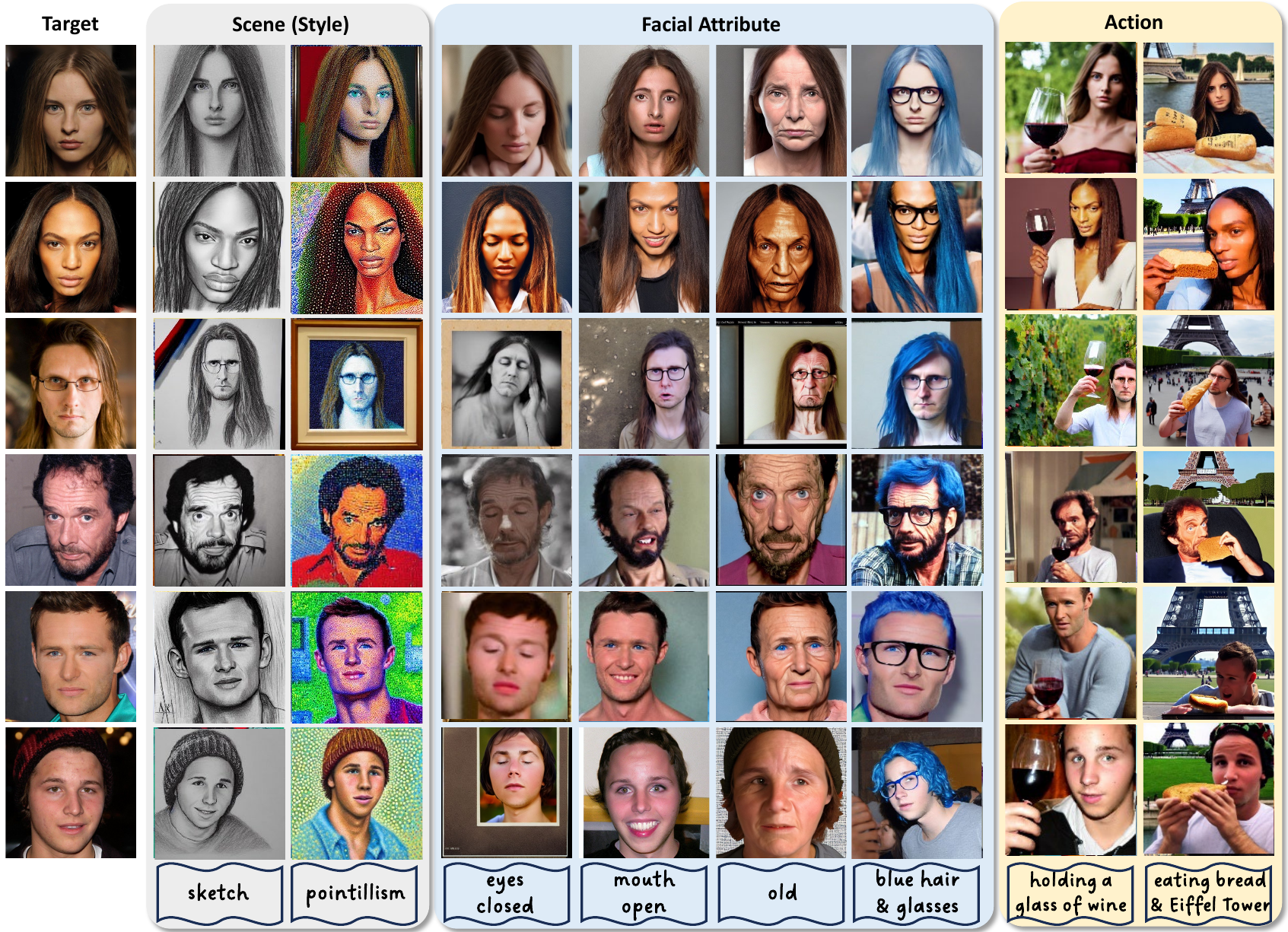}
    \caption{The manipulation diversity of our method, which is shown from various identities, styles, facial attributes, and actions.}
    \label{figure4}
\end{figure*}

\begin{table*}[htb]
        \footnotesize
        \centering
        \caption{Quantitative evaluation between different SOTA methods and ours.}
        \begin{tabular}{cccccccccccccc}
        \toprule[1pt]
        \multicolumn{1}{c}{\multirow{2}*{}}&\multicolumn{4}{c}{Objective Metrics$\uparrow$}& &\multicolumn{3}{c}{User Study$\uparrow$}& &\multicolumn{2}{c}{Params$\downarrow$}&\multicolumn{1}{c}{Time$\downarrow$}\\
        \cline{2-5}\cline{7-9}\cline{11-12}
        ~&Prompt&*ID (F)&*Detect (F)& ID (P)& &Prompt (U)&ID (U)&Quality& &Train&Add&(min)\\
        \midrule[0.05pt]
        DreamBooth~\cite{ruiz2023dreambooth}& 0.249 & 0.488    & 85.2$\%$ & 
0.413 &&0.20&0.07&0.02&& 9.82$\times$$10^{8}$ & 0 &16\\
        Custom Diffusion~\cite{kumari2023multi}& 0.252 & 0.492 & 84.9$\%$ & 0.369 & &0.02&0.16&0.02&& 5.71$\times$$10^{7}$ & 0 &12\\
        Textual Inversion~\cite{gal2022image}& 0.236 & 0.340 & 85.1$\%$ & 0.293 & &0.12&0.07&0.15&& 1536 & 0 &24\\
        Prospect~\cite{zhang2023prospect}& 0.217 & 0.492 & 86.3$\%$ & 0.302 & &0.02&0.22&0.13&& 7680 & 3.1$\times$$10^{7}$ & 18\\
        Celeb Basis~\cite{yuan2023inserting}& 0.242 & 0.412 & 87.1$\%$ & 0.312 & &0.06&0.10&0.16&& 1024 & 6.6$\times$$10^{7}$ &$\sim$5\\
        \cellcolor{gray!15}Ours& \cellcolor{gray!15}\textbf{0.263} & \cellcolor{gray!15}\textbf{0.525} &\cellcolor{gray!15} \textbf{88.8}$\%$ &\cellcolor{gray!15} \textbf{0.428} & \cellcolor{gray!15}&\cellcolor{gray!15}\textbf{0.58}&\cellcolor{gray!15}\textbf{0.38}&\cellcolor{gray!15}\textbf{0.52}&\cellcolor{gray!15}& \cellcolor{gray!15}7680 & \cellcolor{gray!15}3.1$\times$$10^{7}$ &\cellcolor{gray!15}$\sim$5\\
            
        \bottomrule[1pt]
        \end{tabular}
        \label{table1}
    \end{table*}

    \begin{table}
    \centering
    \caption{Quantitative evaluation using previous metrics.}
    \begin{tabular}{ccccc}
        \toprule[1pt]
         &  Prompt & ID & Detect & ID (P) \\
         \midrule[0.05pt]
        DreamBooth~\cite{ruiz2023dreambooth} & 0.253 & 0.261 & 77.5$\%$ & 0.241 \\
        Custom Diffusion~\cite{kumari2023multi} & 0.227 & 0.231 & 83.6$\%$ & 0.243 \\
        Textual Inversion~\cite{gal2022image} & 0.198 & \textbf{0.382} & \textbf{87.5}$\%$ & 0.176 \\
        Prospect~\cite{zhang2023prospect} & 0.209 & 0.372 & 84.2$\%$ & 0.193 \\
        Celeb Basis~\cite{yuan2023inserting} & 0.253 & 0.299 & 82.6$\%$ & 0.184 \\
        \cellcolor{gray!15}Ours & \cellcolor{gray!15}\textbf{0.265} & \cellcolor{gray!15}0.366 & \cellcolor{gray!15}85.3$\%$ & \cellcolor{gray!15}\textbf{0.258} \\
        \bottomrule[1pt]
    \end{tabular}
    \label{id_prompt}
\end{table}

\noindent
\textbf{Baseline Methods.} Our task setting is using only one face image to embed the novel ID into the pre-trained Stable Diffusion Model. Thus, for fair comparisons, we only use a single image for all personalized generation methods, but using enough optimization time for different methods. We select six state-of-the-art works as baseline methods for comparisons from three perspectives: \textit{\textbf{(1) Model Fine-Tuning}}: DreamBooth~\cite{ruiz2023dreambooth} (learns a unique identifier and fine-tunes the diffusion model to learn from a set of images) and Custom Diffusion~\cite{kumari2023multi} (retrieves images with similar captions of the target concept and optimizes the cross-attention module with a modifier token); \textit{\textbf{(2) Token Optimization}}: Textual Inversion~\cite{gal2022image} (learns a pseudo-word for a concept within a limited number of images for optimization), ProSpect~\cite{zhang2023prospect} (expands the textual conditioning space with several per-stage textual token embeddings), and Celeb Basis~\cite{yuan2023inserting} (builds a well-defined face basis module to constrict the face manifold); \textit{\textbf{(3) Tuning Free}}: FastComposer~\cite{xiao2023fastcomposer} (deploys a delayed subject conditioning strategy to achieve tuning-free image generation). 


\noindent
\textbf{Metrics.} We evaluate all the methods from objective metrics, user study, parameter amount, and fine-tuning time. \textit{\textbf{(1) Objective Metrics}}: We select \textit{\textbf{Prompt}} (CLIP alignment score~\cite{saharia2022photorealistic} between text and image), \textit{\textbf{ID}} (ID feature similarity score~\cite{deng2019arcface}), and \textit{\textbf{Detect}} (face detection rate~\cite{deng2019arcface}). However, evaluating the ID without the essence of T2I generation (i.e., Prompt-Image alignment has the highest priority) is inappropriate, and we \textbf{\textcolor{purple}{DISCUSS}} the reasons for this problem in Sec.~\ref{discuss}. Thus, we propose a new metric for face personalized generation which is denoted as \textit{\textbf{ID (P)}}. Specifically, if the CLIP score of this image is lower than the threshold (set as 0.23), then the \textit{\textbf{ID (P)}} score of this image is 0. The threshold is the average CLIP score of these images which get higher scores in user study. To distinguish these ID metrics, we denote \textit{\textbf{*ID (F)}} and \textit{\textbf{*Detect (F)}} for evaluating the images using ``a photo of the face of V*''. \textit{\textbf{(2) User Study}}: We select more than $\sim$20 volunteers and generate $\sim$200 images, to evaluate different methods from \textit{\textbf{Prompt (U)}} (Prompt-Image alignment), \textit{\textbf{ID (U)}} (ID accuracy), and \textit{\textbf{Quality}} (image quality). \textit{\textbf{(3) Parameter Amount}}: We compare the parameter amount from parameters to be learned (\textit{\textbf{Train}}) and the total introduced parameters (\textit{\textbf{Add}}). \textit{\textbf{(4) Time}}: We evaluate the fine-tuning time of different methods to show efficiency performance.

\begin{figure*}[t]
    \centering
    \includegraphics[scale=0.7]{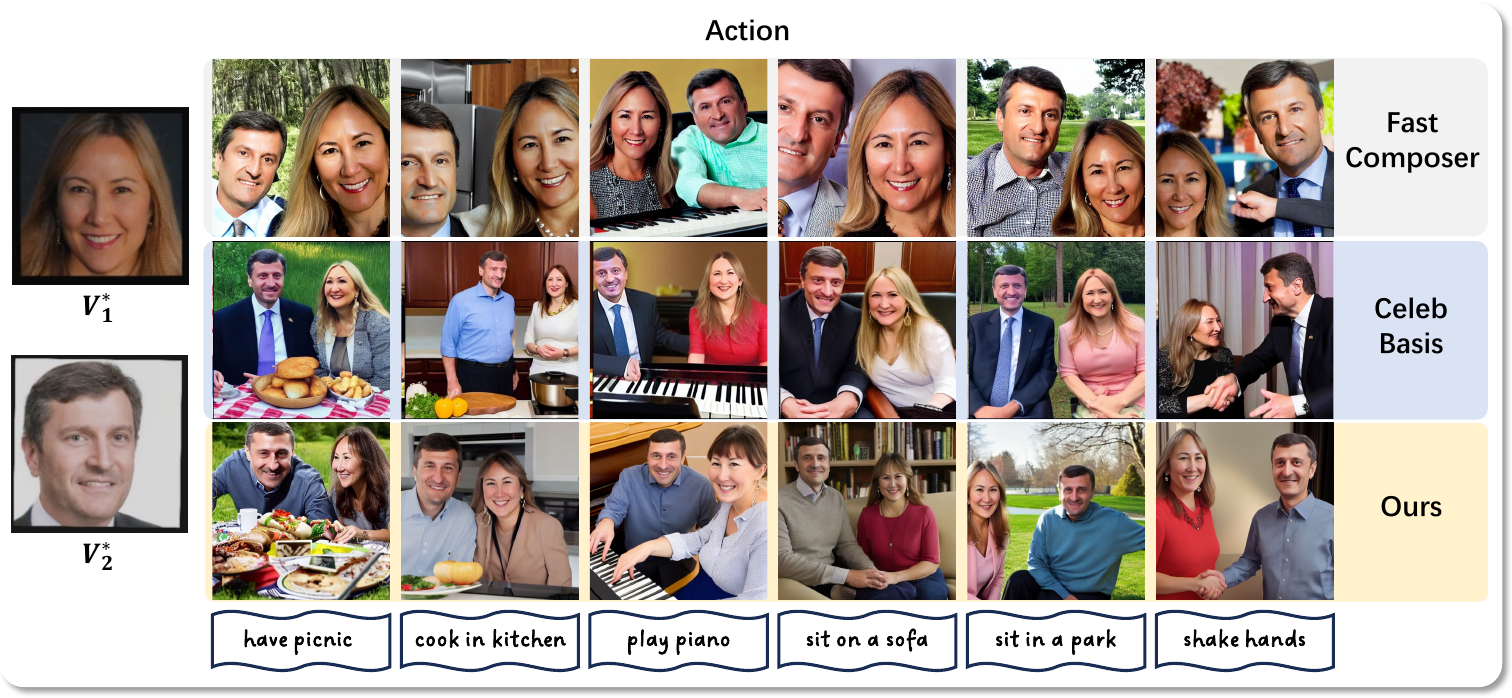}
    \caption{Multi-ID action manipulation comparisons of Celeb Basis~\cite{yuan2023inserting}, FastComposer~\cite{xiao2023fastcomposer}, and ours. FastComposer only focuses on faces and fails to interact with other concepts, such as ``shake hands'', ``sofa'', and ``picnic''. Although Celeb Basis can generate text-aligned images, it shows lower identity preservation.}
    \label{action}
\end{figure*}

\begin{figure}[t]
    \centering
    \includegraphics[scale=0.55]{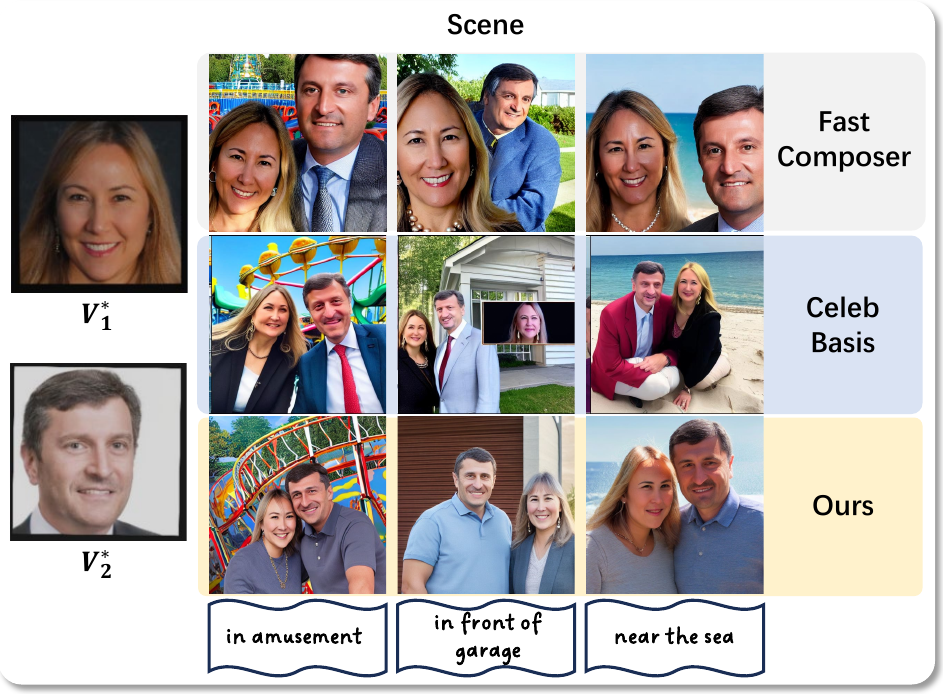}
    \caption{Multi-ID scene manipulation comparisons of Celeb Basis~\cite{yuan2023inserting}, FastComposer~\cite{xiao2023fastcomposer}, and ours. As for FastComposer, the faces of target IDs take over most of the generated picture and some concepts are lost, like ``garage''. As for Celeb Basis, its learned IDs are less precise and may generate artifacts (i.e., a head of a woman which should not exist in the photo).}
    \label{scene}
    \vspace{-0.3cm}
\end{figure}

\begin{figure}[t]
    \centering
    \includegraphics[scale=0.56]{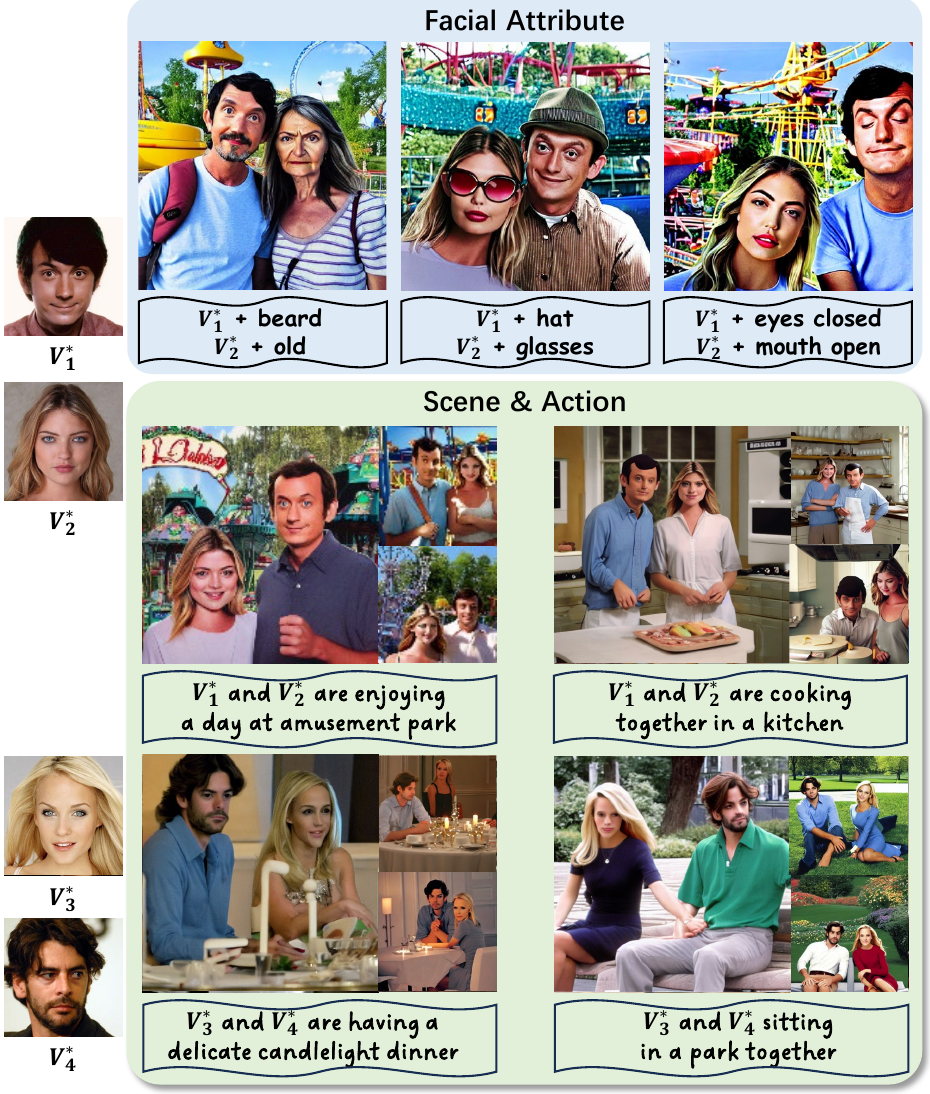}
    \caption{Our multi-ID generation results tested in more complex scenarios, showcasing the diversity of generated images and the ability to interact with complex concepts.}
    \label{multiperson}
    \vspace{-0.1cm}
\end{figure}

\noindent
\subsection{\textcolor{purple}{*DISCUSSION* for ID Similarity Evaluation.}}
\label{discuss}
 As shown in Fig.~\ref{face_generation}, different from face image generation (i.e., using ``a photo of the face of V*'') and editing, achieving Prompt-Image alignment has the highest priority in our task. We have to note this important issue from two perspectives: \textit{\textbf{(1) Explanations for the Previous ID Similarity Metric:}} As shown in Tab.~\ref{id_prompt}, the reason for ID similarity metric less than 0.4, is due to differences in face region resolution. These T2I generated images require face cropping, scaling, and alignment. Consequently, the ID scores are lower than in previous face image generation methods. To fairly evaluate ID similarity under the setting of face photo generation, we conduct ID similarity evaluation using the same-resolution generated face images as the input images with metrics \textbf{\textit{*ID (F)}} and \textbf{\textit{*Detect (F)}}, as shown in Tab.~\ref{table1}. \textit{\textbf{(2) Evaluating ID Considering Text-to-Image Alignment:}} ID evaluation ignoring the T2I alignment shows ``\textbf{fake high}'' ID scores. As shown in Fig.~\ref{teaser} and Tab.~\ref{id_prompt}, we observe that the previous methods failed to generate the images aligned with prompt such as ``a V* is enjoying a cup of latte'' and only generate face images due to attention overfitting, but they had the higher ID scores. This ID evaluation ignores pre-requisite of T2I alignment. Therefore, considering the essence of T2I generation, we have to propose \textbf{ID (P)} for fair comparisons. 

\begin{figure}
    \centering
    \includegraphics[width=\linewidth]{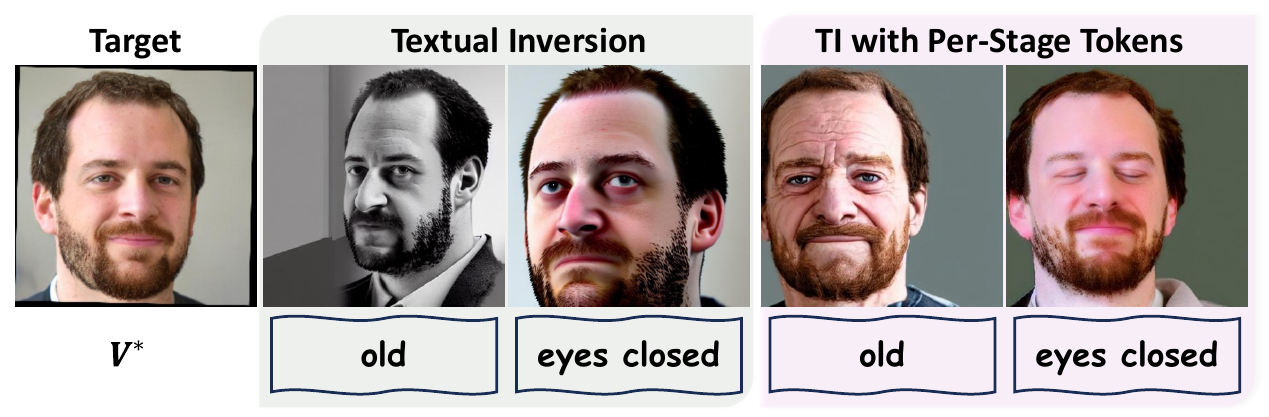}
    \caption{Ablation study of using per-stage tokens with previous methods. The per-stage tokens strategy enables our method to manipulate the facial attributes of target face, and also works for previous methods.}
    \label{per-stage token}
\end{figure}

\subsection{Single ID Embedding and Manipulation} 

We first utilize the same prompts to evaluate five different state-of-the-art methods and ours on single ID embedding and manipulation, as shown in Fig.~\ref{figure3}. We evaluate the performance from three different levels: facial attributes (e.g., age and hairstyle), actions (i.e., human motion and interactions with other objects), and combinations of facial attributes and actions. Textual Inversion~\cite{gal2022image} and Prospect~\cite{zhang2023prospect} tend to overfit the input image, so they fail to interact with other concepts. Although DreamBooth~\cite{ruiz2023dreambooth} and Custom Diffusion~\cite{kumari2023multi} successfully generate the image of interaction of human and concept, the generated identities fail to maintain the ID consistency with the target images. Celeb Basis~\cite{yuan2023inserting} successfully generate the human-object interaction actions, but they fail to manipulate the facial attributes of target identities well. Additional results showcased in Fig.~\ref{figure4} further illustrate the diverse range of manipulations accomplished by our methods in terms of scene (stylization), facial attributes, and action representation within the context of single-person image generation.

\subsection{Quantitative Evaluation} 

As shown in Tab.~\ref{table1}, our method achieves the SOTA performance in the Prompt-Image alignment evaluation and ID (Face) similarity. Due to attention overfit, Textual Inversion~\cite{gal2022image}, Prospect~\cite{zhang2023prospect} show poor Prompt-Image alignment. 
Since that achieving Prompt-Image alignment has the highest priority, we propose a new metric ID (P), which requires the generated images have to achieve the task of semantic-fidelity, and then we calculate their ID scores. Our method achieves better ID (P) scores than the other methods and ours is excellent in Prompt-Image alignment evaluation. This improvement is from two reasons: (1) Attention Overfit Alleviation: our face-wise attention loss is able to alleviate the attention overfit problem of previous methods such as DreamBooth~\cite{ruiz2023dreambooth}, Prospect~\cite{zhang2023prospect}, and Texutal Inversion~\cite{gal2022image}. Our method can make the ID embedding focus on the face region, instead of the whole image. (2) Attribute-Aware Tokens: Compared to Celeb Basis~\cite{yuan2023inserting}, our method does not introduce too much face prior and represents one ID as five per-stage tokens, which can balance the trade-off between ID accuracy and manipulation ability. Our expended textual conditioning space has a strong disentanglement and control ability of attributes (e.g., action-related objects and facial attributes) than Celeb Basis~\cite{yuan2023inserting}.

\begin{figure}[t]
    \centering
    \includegraphics[scale=0.65]{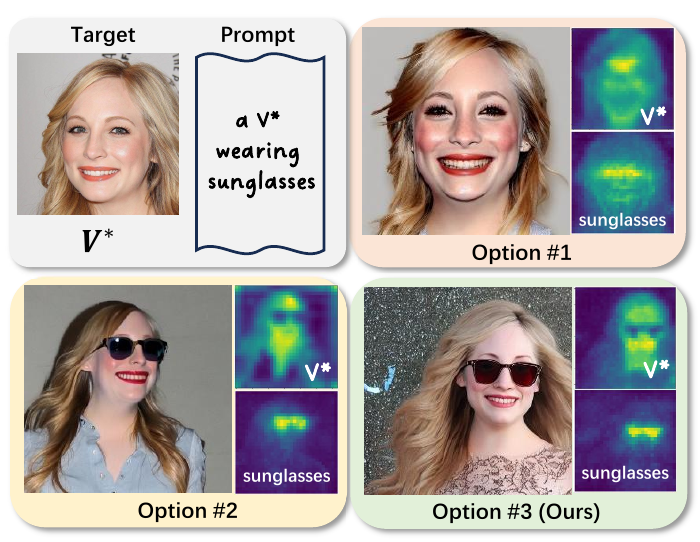}
    \caption{Ablation study of different options for attention loss. Option $\#1$ inferences other concepts, and option $\#2$ still disrupts regions like the corners of the activation map for $V^*$ that beyond its scope.}
    \label{ablation_attention_loss}
\end{figure}

\subsection{Multi-ID Embedding and Manipulation}

As shown in Fig.~\ref{action} and Fig.~\ref{scene}, we illustrate the circumstances where two IDs appear in the same scene and some interactive actions between them. Though Celeb Basis~\cite{yuan2023inserting} can achieve competitive prompt alignment as ours, the generated identity is less precise which leads to their poor identity similarity as shown in Tab.~\ref{table1}. We hypothesis that in the absence of explicit regularization, the learned ID embedding may be sub-optimal, as they still can not disentangle the identity representation from the other latent factors. For instance, the results in Fig.~\ref{action} suggest that their learned ID embedding not only focuses on identity but also incorporates additional information, such as clothing (e.g., the consistent presence of a suited man). In the experiments compared with FastComposer~\cite{xiao2023fastcomposer}, The generated images by FastComposer predominantly feature the faces of the target IDs, occupying a significant portion of the images and it seems like that the characters are directly pasted into the picture, resulting in a disharmonious appearance. Besides, it is difficult for FastComposer to interact with other concepts (like ``picnic'' and ``garage'') and generate the correct action (like ``sitting'', ``shaking'', and ``cooking'') because of the aforementioned semantic prior forgetting problem.  As shown in Fig.~\ref{multiperson}, we experiment on more complex scenarios in multi-ID generation, which showcases the high generation diversity and good interactive ability of our method.


\begin{figure}[t]
    \centering
    \includegraphics[scale=0.46]{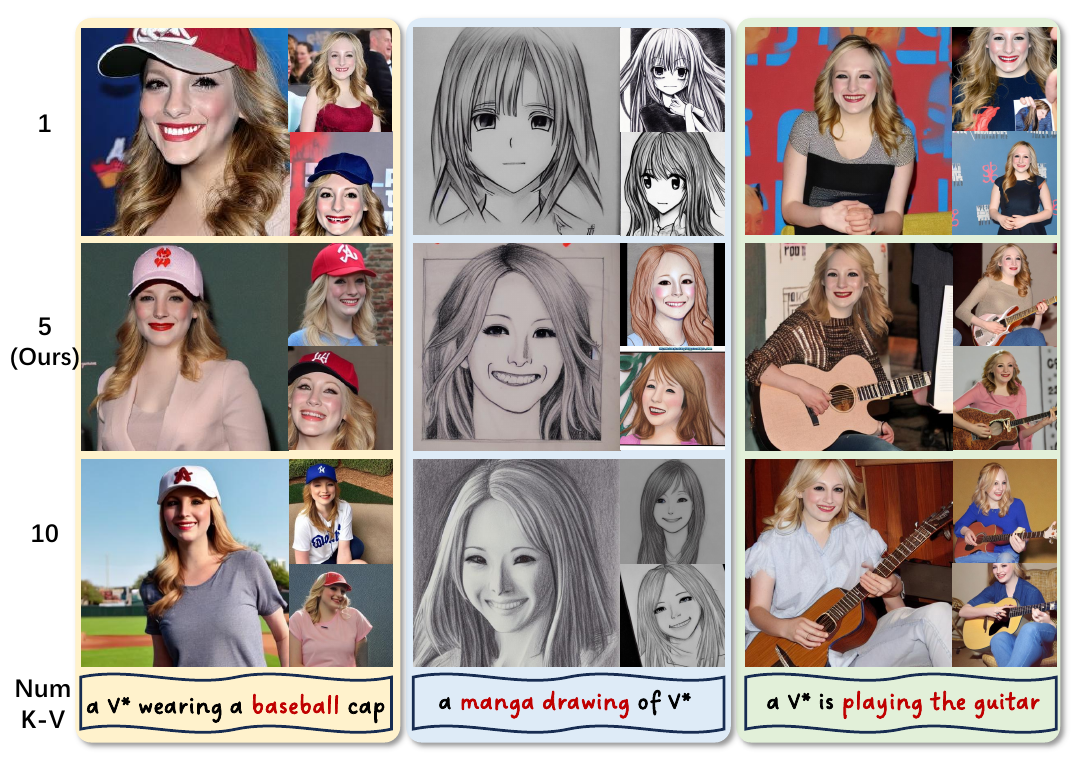}
    \caption{Ablation study of utilizing different number of K-V pairs. Using only 1 K-V pair can not sufficient maintain the ID features. And adopting too many K-V pairs would not bring significant improvements to generation quality. Thus, we finally select 5 K-V pairs.}
    \label{ablation_kv_split}
\end{figure}

\subsection{User Study} 

To make our results more convincing and incorporate a broader range of user perspectives, we further conduct a user study, which can be found in Tab.~\ref{table1}. Our method obtains better preference than previous work among the participating users, including better Prompt-Image alignment, ID similarity to the target reference image, and image quality. This shows that our semantic-fidelity embedding can enable better interactive generation ability and is potential to exploit the powerful manipulation capabilities of the Stable Diffusion Model itself.

\subsection{Efficiency Evaluation} 

As shown in Tab.~\ref{table1}, we have advantages in introduced parameter amount and fine-tuning time. Celeb Basis~\cite{yuan2023inserting} introduces a basis module and a pre-trained face recognition model, but these are large optimization burdens and a too strong facial prior can disrupt the interaction between faces and other concepts. We utilize the prior from the T2I model itself, reducing the introduction of additional parameters and further enhancing the facial manipulation ability of T2I models.

\begin{table}[t]
    
    \setlength\tabcolsep{11pt}
    \centering
    \caption{Quantitative evaluation of ablation study.}
   
    \begin{tabular}{ccccc}
    \toprule[1pt]
    ~ & Prompt$\uparrow$ & ID$\uparrow$ & Detect$\uparrow$ & ID (P)$\uparrow$ \\
    \midrule[0.05pt]
    5 K-V \& \#1 & 0.201 & 0.382 & 85.9$\%$ & 0.190 \\
    5 K-V \& \#2 & 0.246 & 0.324 & 86.4$\%$ & 0.248 \\
    1 K-V \& \#3 & 0.205 & 0.323 & \textbf{88.2}$\%$ & 0.203 \\ 
    10 K-V \& \#3 & \textbf{0.265} & 0.337 & 85.9$\%$ & 0.251 \\
    \cellcolor{gray!15}{5 K-V \& \#3} & \cellcolor{gray!15}\textbf{0.265} & \cellcolor{gray!15}\textbf{0.366} & \cellcolor{gray!15}85.3$\%$ & \cellcolor{gray!15}\textbf{0.258} \\
    \bottomrule[1pt]
    \end{tabular}
    \label{ablation}
\end{table}

\subsection{Ablation Study}
\label{subsec:ablation}
\noindent
\textbf{Different Effects of $\bm{P_{i}^{K}}$ and $\bm{P_{i}^{V}}$ Tokens.} The semantic information of per-stage tokens is important for the interpretation of diffusion-based generation process, especially for the different effects of $\bm{P_{i}^{K}}$ and $\bm{P_{i}^{V}}$ tokens. As shown in Fig.~\ref{per-stage token} we add experiments of using the Per-Stage Token Optimization with previous Textual Inversion, which shows its fine-grained control ability, such as the manipulation of facial attributes. To further investigate this, as shown in Fig.~\ref{kv_interpolation}, we thoroughly explore $\bm{P_{i}^{K}}$ and $\bm{P_{i}^{V}}$ tokens from two perspectives: (1) Progressively Adding: We add different ${\{(\bm{P_i^{K}}, \bm{P_i^{V})}\}}_{1\leq i \leq 5}$ tokens to the conditioning information in ten steps. We found that the initial tokens influence more the layout of generation content (e.g., face region location, and poses), while the latter tokens effect more the ID-related details. (2) Progressively Substituting: We then substitute different $\bm{P_{i}^{K}}$ and $\bm{P_{i}^{V}}$ tokens of ${\{(\bm{P_i^{K}}, \bm{P_i^{V})}\}}_{1\leq i \leq 5}$. We found that $\bm{P_{i}^{V}}$ contribute to the vast majority of ID-related information, and the $\bm{P_{i}^{K}}$ contribute more to textual details, such as environment lighting.

\begin{figure}
    \centering
    \includegraphics[scale=0.6]{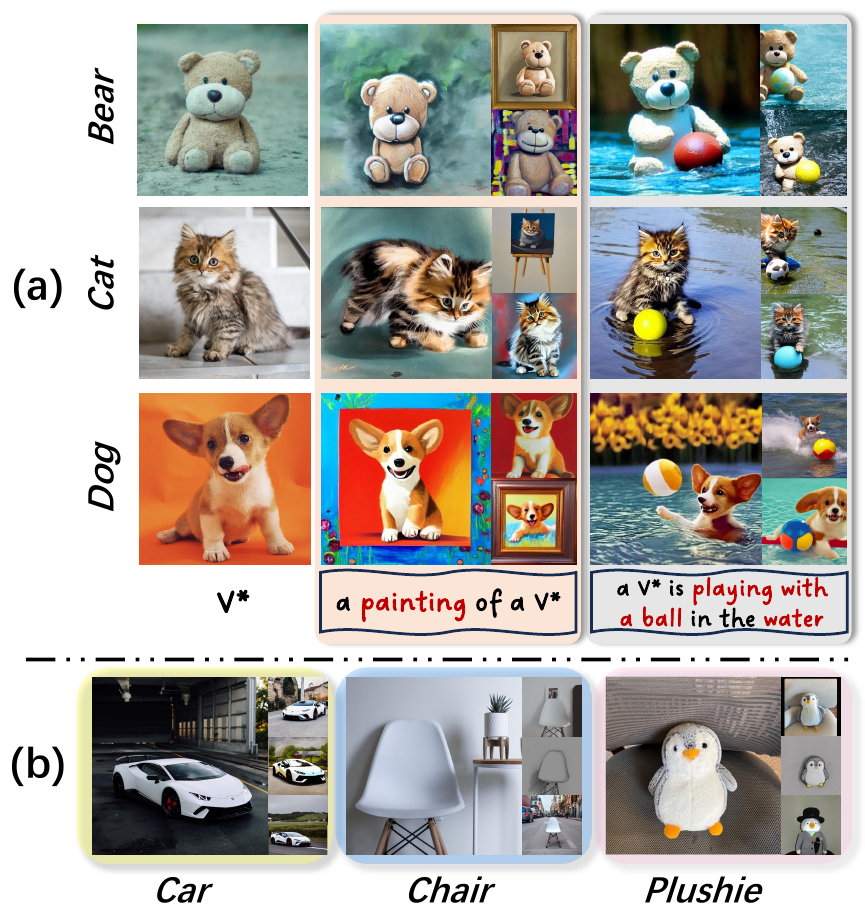}
    \caption{Using our ID embedding method for non-face concepts. In each block of part \textbf{(b)}, the target object is displayed on the left, while on the right, from top to bottom, are the images labeled as ``a photo of V*'', ``stylization of V*'', and ``V* under different scenes''.}
    \label{more objects}
\end{figure}

\begin{figure*}[t]
    \centering
    \includegraphics[width=1.0\linewidth]{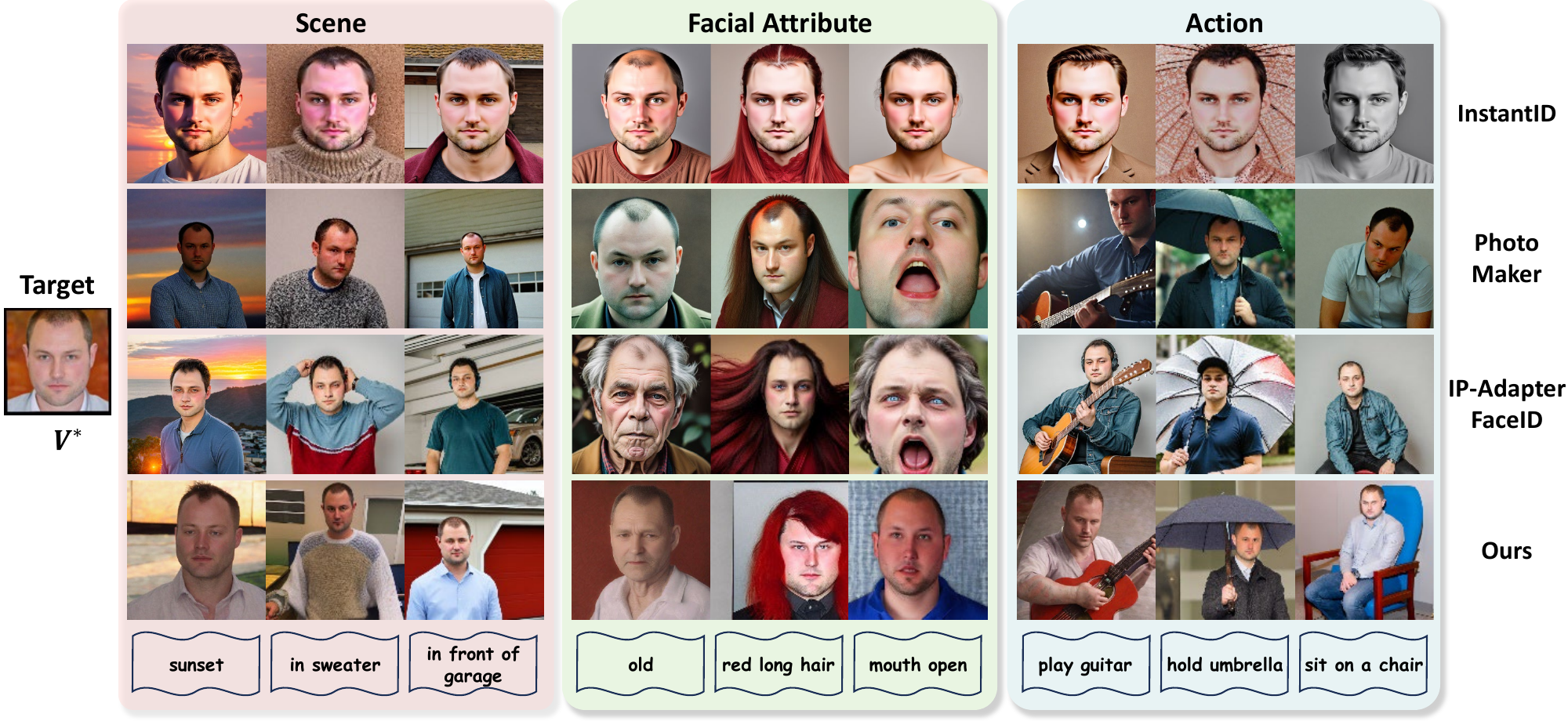}
    \caption{Using our ID embedding method for Stable Diffusion XL. InstantID~\cite{wang2024instantid} tends to generate a face photo of target ID. PhotoMaker~\cite{li2023photomaker} and IP-Adapter-FaceID~\cite{ye2023ip} can not achieve fine-grained text guided facial attribute controlling. Our method can achieve better interactive generation with the other concepts (e.g., chair) than the other methods.}
    \label{sdxl}
\end{figure*}

\noindent
\textbf{Attention Loss.} We thoroughly investigate three options for face-wise attention loss. The option $\#1$ only regularizes on the $V^*$ token and the option $\#2$ regularizes the prompt-length tokens. As shown in Fig.~\ref{ablation_attention_loss}, option $\#1$ affects the other concept embeddings in the T2I model, which results in non-ID concepts cannot be generated, such as sunglasses. Although the option $\#2$ can reduce the influence of too much ID attention, the activation region of $V^*$ still disrupts regions beyond its scope, which can be seen in the corners of the feature activation map for $V^*$. Our final adopted option is $\#3$, which calculates the attention loss among the whole text attention maps generated by each token. This option prevents the learned token from overfitting to other regions and only focus on the face region. Drawn from Tab.~\ref{ablation}, as more tokens are token into the attention loss regularization, the prompt score rises. We think the reason lies in two perspective: (1) The regularization on the $V^*$ token ensures it to focus on face region and prevents it from disturb the other concepts; (2) The regularization applied to all other tokens serves as an additional penalty, preserving their ability to implicitly disentangle the $V^*$ token from the rest of the tokens. Our loss strategy only addresses the attention overfitting, improving the ID accuracy and interactivity with other concepts, but the manipulation capacity for the high text2image alignment and diversity still needs to be improved by us and other diffusion-based generative model researchers.

\noindent
\textbf{The Number of K-V Feature Pairs.} As shown in Fig.~\ref{ablation_kv_split}, we explore the influence of K-V pair numbers. When using only one pair of K-V, the learned ID-related tokens fail to maintain good ID accuracy and interact with other complex concepts and attributes. However, adopting too many K-V pairs (e.g., 10 pairs) fails to bring significant improvements of diversity or quality, and this is no doubt a huge computational burden. In our method, we select 5 K-V pairs, which balance the trade-off of representing capacity and computation. As shown in Tab.~\ref{ablation}, the Prompt and identity scores of setting 1 K-V with option $\#3$ are lower than 5 K-V with option $\#3$ and 10 K-V with option $\# 3$. While the 10 K-V with option $\#3$ shows the same prompt score compared to 5 K-V with option $\#3$, it exhibits lower identity similarity.

\subsection{Generalization}

\noindent
\textbf{Embedding Other Objects.} We further validate our methods on other objects. Compared to Celeb Basis~\cite{yuan2023inserting}, our method does not introduce face prior from other models (e.g., a pre-trained face recognition model or basis module). As shown in Fig.~\ref{more objects}, we adopt animals (\textbf{Bear}, \textbf{Cat}, and \textbf{Dog}) and general objects (\textbf{Car}, \textbf{Chair} and \textbf{Plushie}) for experiments, which show the generalizability of our method.

\noindent
\textbf{Using Stable Diffusion XL.} To validate the generalization to the latest version of Stable Diffusion Model, we select SDXL model~\cite{podell2023sdxl} \texttt{stable-diffusion-xl-base-1.0} as the target model and the newly released methods using it for comparisons. As shown in Fig.~\ref{sdxl}, we compare with the SOTA methods InstantID~\cite{wang2024instantid}, PhotoMaker~\cite{li2023photomaker}, and IP-Adapter-FaceID~\cite{ye2023ip}. InstantID~\cite{wang2024instantid} can only generate the face photo and fails to manipulate other actions or facial attributes. Although PhotoMaker~\cite{li2023photomaker} and IP-Adapter-FaceID~\cite{ye2023ip} could generate the target ID under different scenes and actions, they can not handle complex actions (e.g., ``sit on a chair'') and accurate facial attribute controlling. Additionally, IP-Adapter-FaceID~\cite{ye2023ip} even loses the identity information of target person when combined with facial attribute prompts. As shown in the shortcomings of other methods, we found that incorporating additional features into the SDXL model would compromise the semantic-fidelity ability of T2I models, resulting in generated images that are misaligned with the given prompts. In contrast, our approach focuses on learning interactive ID embeddings with diffusion prior itself, which would not disrupt the original semantic understanding capability of the adopted models.

\section{Conclusion}
\label{sec:conclusion}
We propose two novel problem-orient techniques to enhance the accuracy and interactivity of the ID embeddings for semantic-fidelity personalized diffusion-based generation. We analyze the attention overfit problem and propose Face-Wise Attention Loss. This improves the ID accuracy and facilitates the effective interactions between this ID embedding and other concepts (e.g., scenes, facial attributes, and actions). Then, we optimize one ID embedding as multiple per-stage tokens, which further expands the textual conditioning space with semantic-fidelity control ability. Extensive experiments validate our better ID accuracy and manipulation ability than previous methods, and we thoroughly conduct ablation study to validate the effectiveness of our methods. Moreover, our embedding method does not rely on any prior facial knowledge, which is potential to be applied to other categories.

\noindent
\textbf{Ethical Statement.} Our research endeavors are dedicated to addressing specific challenges within multi-modal generation with the overarching aim of advancing the technological landscape within our community. We staunchly oppose any misuse of our technology, such as the unauthorized use of their identity information. To mitigate such risks, we are actively developing watermarking techniques to prevent the misuse of Artificial Intelligence Generated Content.




{
    \small
    \bibliographystyle{IEEEtran}
    \bibliography{bare_jrnl_new_sample4}
}

\vspace{-33pt}
\begin{IEEEbiography}[{\includegraphics[width=1in,height=1.25in,clip,keepaspectratio]{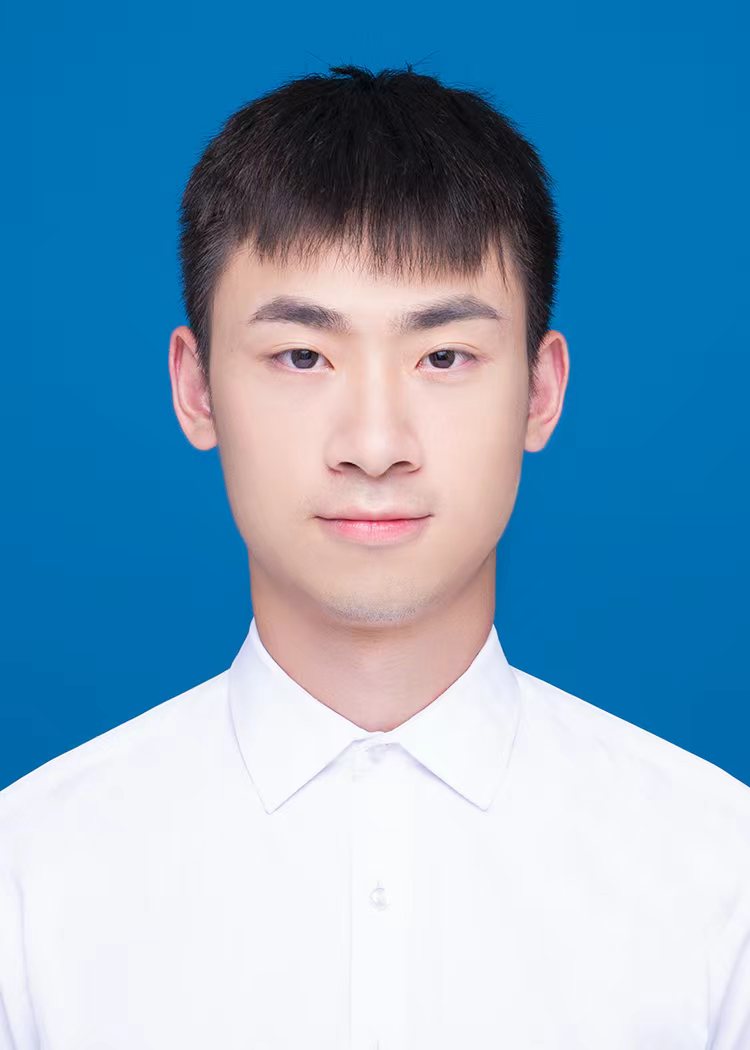}}]{Yang Li} received a BEng degree in Harbin Institute of Technology in 2022. He is currently a master degree candidate in the University of Chinese Academy of Sciences and also at the Institute of Automation, Chinese Academy of Sciences. His research interests are in computer vision and generative models.
\end{IEEEbiography}
\vspace{11pt}

\vspace{-33pt}
\begin{IEEEbiography}[{\includegraphics[width=1in,height=1.25in,clip,keepaspectratio]{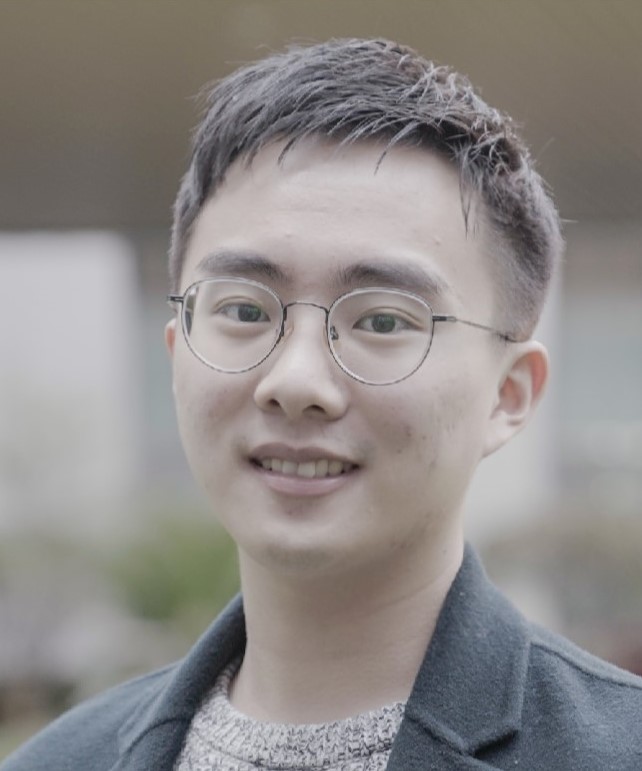}}]{Songlin Yang} received a BEng degree in Nanjing University of Aeronautics and Astronautics in 2021. He is currently a master degree candidate in the University of Chinese Academy of Sciences and also at the Institute of Automation, Chinese Academy of Sciences. His research interests are in computer vision, computer graphics, and machine learning.
\end{IEEEbiography}
\vspace{11pt}

\vspace{-33pt}
\begin{IEEEbiography}[{\includegraphics[width=1in,height=1.25in,clip,keepaspectratio]{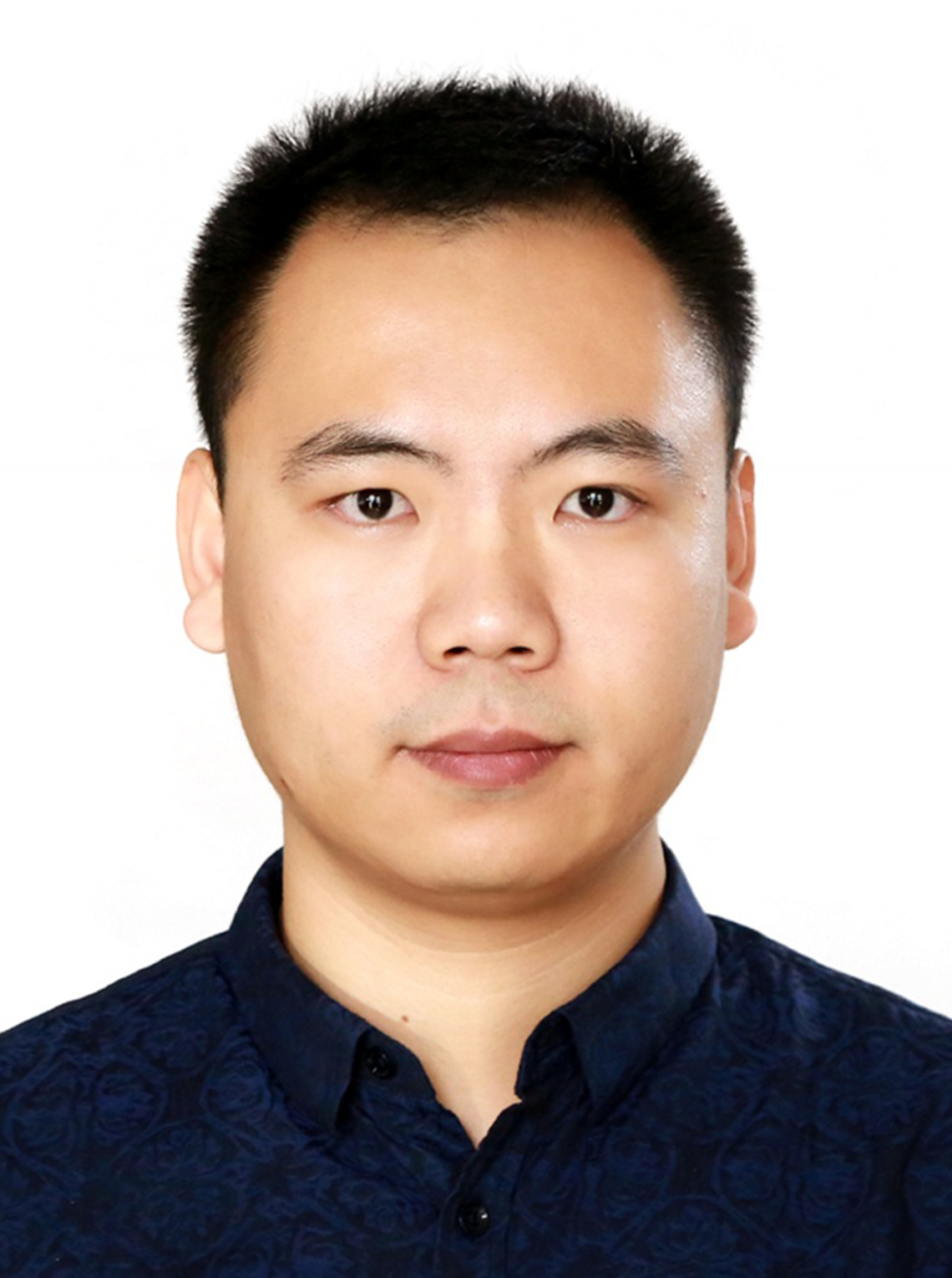}}]{Wei Wang} received his Ph.D. degree in pattern recognition and intelligent systems from the Institute of Automation, Chinese Academy of Sciences (CASIA) in 2012. He is currently an Associate Professor with the State Key Laboratory of Multimodal Artificial Intelligence Systems (MAIS), CASIA. His research interests include artificial intelligence safety and multimedia forensics.
\end{IEEEbiography}
\vspace{11pt}


\vspace{-33pt}
\begin{IEEEbiography}[{\includegraphics[width=1in,height=1.25in,clip,keepaspectratio]{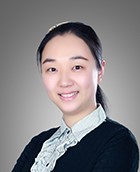}}]{Jing Dong} recieved her Ph.D in Pattern Recognition from the Institute of Automation, Chinese Academy of Sciences in 2010.  Then she  joined the National Laboratory of Pattern Recognition (NLPR) and she is currently Professor in the State Key Laboratory of Multimodal Artificial Intelligence Systems. Her research interests are towards Pattern
Recognition, Image Processing and Digital Image Forensics including digital watermarking, steganalysis and tampering detection. She is a senior member of IEEE. She also has served as the deputy general of Chinese Association for Artificial Intelligence.
\end{IEEEbiography}
\vspace{11pt}

\vfill

\end{document}